\documentclass{article}

\usepackage[utf8]{inputenc}    % input encoding
\usepackage[T1]{fontenc}       % font encoding
\usepackage{CJKutf8}           % suporte a caracteres CJK
\usepackage{amsmath, amssymb, amsfonts}
\usepackage{xfrac}
\usepackage{microtype}         % melhorias de tipografia

\usepackage{booktabs}          % tabelas de qualidade profissional
\usepackage{multirow}          % mesclar células verticalmente
\usepackage{makecell}          % células com múltiplas linhas
\usepackage{threeparttable}    % tabelas com notas de rodapé
\usepackage{tablefootnote}     % notas de rodapé em tabelas
\usepackage{tabularx}          % tabelas com ajuste automático de largura

\usepackage{graphicx}

\usepackage{xcolor}
\usepackage{textcomp}

\usepackage[colorlinks=true, allcolors=blue]{hyperref}
\usepackage{url}
\usepackage{doi}
\usepackage{natbib}
\bibpunct{[}{]}{;}{a}{,}{,}     % formatação de citações do natbib

\usepackage{arxiv}

\title{The Paradox of Poetic Intent in Back-Translation: Evaluating the Quality of Large Language Models in Chinese Translation}

\author{ \href{https://orcid.org/0000-0003-1826-1850}{\includegraphics[scale=0.06]{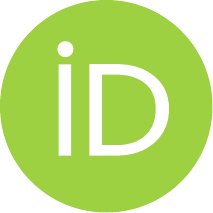}\hspace{1mm}Li Weigang} \\
	Computer Science Department\\
	University of  Brasilia\\
        Brasília, Brazil \\
	\texttt{weigang@unb.br} \\
	\And
	\href{https://orcid.org/0000-0002-1288-7695}{\includegraphics[scale=0.06]{orcid.pdf}\hspace{1mm}Pedro Carvalho Brom} \\
	Math Department\\
	Federal Institute of Brasilia\\
	Brasília, Brazil \\
	\texttt{pedro.brom@ifb.edu.br} \\
}

\renewcommand{\shorttitle}{Poetic Paradox in LLM Back-Translation}

\hypersetup{
pdftitle={The Paradox of Poetic Intent in Back-Translation: Evaluating the Quality of Large Language Models in Chinese Translation},
pdfsubject={cs.CL},
pdfauthor={Li Weigang, Pedro Carvalho Brom},
pdfkeywords={Back-translation, BLEU, CNLP, quasi-self-awareness, paradox of poetic intent, verbatim back-translation},
}

\begin{document}
\maketitle

\begin{abstract}
	The rapid advancement of large language models (LLMs) has reshaped the landscape of machine translation, yet challenges persist in preserving poetic intent, cultural heritage and handling specialized terminology in Chinese-English translation. This study constructs a diverse corpus encompassing Chinese scientific terminology (e.g., CNKI abstracts), historical translation paradoxes (e.g., the Xue Dejiong dilemma) and literary metaphors (e.g., the spatio-temporal narrative of Daolang’s Hua Yao). Utilizing a back-translation and Friedman test-based system, we evaluate four metrics, BLEU, CHRF, TER and Semantic Similarity, across six major LLM platforms (e.g., GPT-4.5, DeepSeek V3) and three traditional translation tools through Chinese→English→Chinese back-translation experiments. Key findings include: 1) Back-translation of scientific abstracts outperforms direct translation, suggesting alignment optimization during the process, while traditional tools retain advantages in linguistically distinctive texts (e.g., Sogou Translate BLEU 0.57); 2) LLMs exhibit inconsistent performance on cultural and literary texts, revealing the ``paradox of poetic intent'', pre-trained models often sacrifice deeper meaning for literal consistency, with DeepSeek V3 excelling in metaphor retention (mean BLEU 0.60, variance 0.10); 3) Some LLMs display ``verbatim back-translation,'' demonstrating data-driven ``quasi-self-awareness'' and potential beyond conventional methods; 4) Addressing limitations of character-based BLEU, we develop a Jieba-segmentation-based BLEU method integrating word frequency and n-gram weighting, offering a new perspective for Chinese translation evaluation. This study dissects bottlenecks in Chinese-English translation, objectively assessing the Chinese translation capabilities of LLMs and tools and providing insights for optimizing CNLP tasks and enhancing cultural sensitivity.
\end{abstract}

% keywords can be removed

\section{INTRODUCTION}

The synergistic advancement of Large Language Models (LLMs) and artificial intelligence has ushered in a paradigm shift in natural language processing (NLP) \cite{vaswani2017attention}. Transformer-based architectures have enabled remarkable progress in text generation and cross-lingual translation. Nevertheless, applying LLMs to non-English languages, especially structurally and culturally distinct systems like Chinese, remains an enduring challenge. Key difficulties include the preservation of poetic intent, cultural nuance and domain-specific terminology \cite{zhang2021study, zhong2024beyond}.

Spoken by over 1.6 billion people worldwide \cite{eberhard2023ethnologue}, Chinese presents unique linguistic complexities. Its syntactic ambiguity, idiomatic richness and specialized terminology (e.g., heterocyclic compounds like ``thiazole'' and ``pyridine'') have long hindered machine understanding. Chinese$\Leftrightarrow$English translation, in particular, continues to be one of the most persistent bottlenecks in multilingual AI systems \cite{sun2021chinesebert,cao2020zero}. Compounding the issue is the multimodal nature of Chinese characters, rooted in the ancient Liu Shu (Six-Writings) system and exemplified in Xu Shen’s Shuowen Jiezi \cite{zhou2014six}, which combine ideographic and phonosemantic features. These historical mechanisms offer valuable insight into modern grapheme–phoneme encoding strategies employed by LLMs \cite{weigang2024six}.

In today's globalized landscape, cross-lingual translation serves as infrastructure across international trade, scientific communication and cultural exchange. In 2024, China’s foreign trade volume approached USD 6 trillion\footnote{www.chinadaily.com.cn/a/202502/12/WS67abd994a310a2ab06eab9e6.html}, where translation quality directly affects contract execution, product compliance and supply chain performance. Scientifically, more than 25\% of the world's SCI publications in 2023 originated from China, with over 40\% requiring English translation \cite{jiang2020nsfc}. Culturally, the 2025 global release of Nezha: Rebirth of the Demon Child 2 earned over USD 200 million at the North American box office\footnote{www.chinadaily.com.cn/a/202503/04/WS67c5d351a310c240449d8511.html}, highlighting the critical role of dubbing and subtitling in international media.

Despite their success in English-centric tasks, LLMs exhibit structural limitations when applied to Chinese. Transformer models such as ChatGPT are primarily trained on word-based English tokens using long-range contextual attention \cite{vaswani2017attention}, well-suited to space-delimited languages. In contrast, Chinese poses two fundamental challenges \cite{wong2022introduction,weigang2025threshold}: (1) the lack of explicit word boundaries, making tokenization ambiguous and (2) a more uniform information distribution at the character level, which dominates current training and evaluation practices.

However, modeling Chinese purely at the character level is suboptimal. 
While one- and two-character words constitute approximately 95\% of high-frequency items 
(e.g., ``\begin{CJK*}{UTF8}{gbsn}技术\end{CJK*}'' [technology], 
``\begin{CJK*}{UTF8}{gbsn}翻译\end{CJK*}'' [translation]) \cite{liu1986chinese}, 
many semantically dense three- and four-character expressions 
(e.g., ``\begin{CJK*}{UTF8}{gbsn}机器人\end{CJK*}'' [robot], 
``\begin{CJK*}{UTF8}{gbsn}人工智能\end{CJK*}'' [AI] 
and idioms like ``\begin{CJK*}{UTF8}{gbsn}刻舟求剑\end{CJK*}'') 
carry disproportionate contextual weight. 
Treating characters as isolated units often leads LLMs to lose the semantic integrity of these word-level constructs, resulting in unnatural phrasing or cultural misalignment. 
This phenomenon helps explain the comparative underperformance of LLMs on Chinese tasks, particularly those involving metaphor, poetry or specialized jargon.

Empirical evaluations reflect this gap. Mainstream machine translation systems report BLEU scores \cite{papineni2002bleu} below 0.45 for English-to-Chinese tasks and under 0.27 for Chinese-to-English, substantially lower than for many other language pairs (e.g., Russian–English BLEU 43.53) \cite{zhu2025overcoming}. This discrepancy is especially evident in sensitive domains such as science, law and healthcare. Although hybrid architectures, such as sparse Mixture-of-Experts (MoE) models integrated into GPT and LLaMA, show promise \cite{zhu2025overcoming}, emerging platforms (e.g., CroK, DeepSeek, Gemini) lack rigorous benchmarking, especially for content involving cultural depth, historical references or scientific terminology.

\begin{figure}[htbp]
\centering
\includegraphics[width=0.6\textwidth]{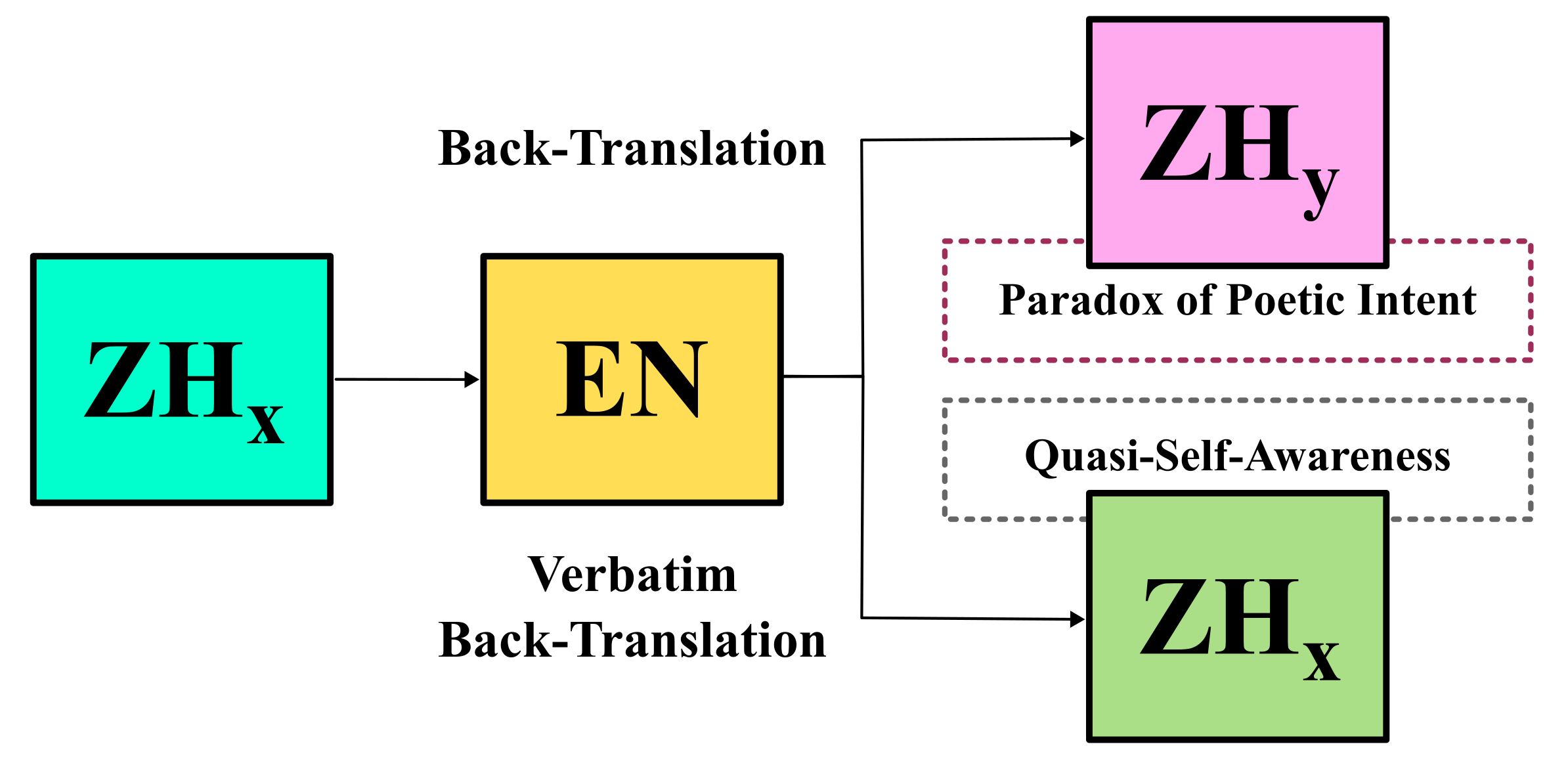}
\caption{Conceptual Diagram: The Paradox of Poetic Intent in back-translation vs. emergent Quasi-Self-Awareness in LLM verbatim back-translation, where $ZHx$ is the original Chinese text, $EN$ is the translated English text and $ZHy$ is the back-translated Chinese text.}
\label{imagery_vs_selfawareness}
\end{figure}

To address these challenges, comprehensive innovation is needed in segmentation strategies, tokenization design and corpus development to establish a robust Chinese NLP ecosystem. Against this backdrop, this study introduces a Chinese $\rightarrow$ English $\rightarrow$ Chinese back-translation framework to evaluate the semantic fidelity and cultural retention of LLM outputs \cite{edunov2018understanding, artetxe2018unsupervised}. Central to this is the conceptual framework we term the ``Paradox of Poetic Intent'', a tension between surface-level fidelity and deeper cultural meaning during LLM-mediated translation. Figure~\ref{imagery_vs_selfawareness} illustrates this tension in relation to emergent LLM behaviors such as verbatim memory reconstruction.

Verbatim back-translation may be viewed as a form of internal memory mapping within LLMs. Recent studies have enriched our understanding of LLM memory: Brown et al. (2020) \cite{brown2020language} introduced parametric memory encoded within model weights; Shan et al. (2025) \cite{shan2025cognitive} categorized memory types from a cognitive standpoint; Modarressi et al. (2024) \cite{modarressi2024memllm} proposed read–write modules for rare event retention; Yang et al. (2024) \cite{yang2024memory3} externalized memory to reduce model size and inference cost; and Zhang et al. (2024) \cite{zhang2024survey} provided a systematic survey of memory in LLM agents.

These works collectively inform our understanding of verbatim back-translation as a manifestation of latent memory. Building on this, we argue that LLMs demonstrate a form of quasi-self-awareness: an emergent capacity to preserve stylistic and contextual coherence without explicit instruction. This capacity opens new conceptual territory for exploring the tension between linguistic accuracy and deeper meaning, embodied by the Paradox of Poetic Intent.

To investigate this, we construct a domain-diverse dataset comprising: (1) historical paradoxes in scientific naming (e.g., the Xue Dejiong’s dilemma \cite{hejuan2019nomencl,weigang2025xdj}); (2) metaphoric and cultural texts (e.g., Daolang’s Hua Yao and its spatiotemporal narrative); and (3) idiomatic and terminological expressions from 295 CNKI academic abstracts. Using both single-sample and multi-sample paradigms, we evaluate six major LLMs (e.g., GPT-4.5, Claude 3.7, DeepSeek V3, Grok 3, Gemini 2.0) and three commercial translation platforms under the back-translation framework, employing statistical significance tests to validate our findings.

Recent studies such as \cite{chen2024benchmarking} and \cite{zhao2025fuxi} have explored the capabilities of LLMs in translating and generating Classical Chinese texts, with a particular focus on evaluating adequacy, fluency, and elegance. While their contributions provide valuable insights into translation performance on poetic or ancient literary corpora, our study shifts the emphasis toward the cognitive and generative mechanisms at play when LLMs encounter culturally embedded and semantically intricate expressions. Specifically, we examine how LLMs behave under verbatim back-translation tasks, and uncover emergent phenomena such as the Paradox of Poetic Intent that reveal the boundaries of alignment, stylistic fidelity, and latent semantic dissonance in cross-lingual contexts.

\textbf{The main contributions of this study are as follows:}
\begin{itemize}
\item A back-translation framework integrated with multi-sample Friedman testing is developed to assess model performance across BLEU, CHRF, TER and Semantic Similarity. Results confirm the ``Paradox of Poetic Intent'': LLMs tend to compromise literary depth in favor of literal alignment. Claude 3.7 and DeepSeek V3 demonstrated superior metaphor preservation (average BLEU of 0.60, variance 0.10).

\item Back-translations of scientific abstracts outperformed direct translations in adequacy, suggesting iterative translation may enhance semantic alignment. Conversely, commercial tools retained comparative advantages in language-specific scenarios, for example, Sogou Translate yielded a BLEU of 0.57 on scientific content.

\item ``Verbatim back-translation'' effects were frequently observed in LLMs like ChatGPT and DeepSeek, indicating quasi-self-awareness\footnote{In this context, `quasi-self-awareness refers to the empirical observation that for certain inputs \(ZH_x\), the back-translated output \(ZH_y\) is nearly identical.}, where the model implicitly reconstructs prior inputs without explicit prompts. This emergent behavior hints at deeper cognitive modeling capacities.

\item A novel BLEU evaluation method incorporating Jieba segmentation, word frequency and n-gram weighting is proposed to reflect Chinese-specific features. This approach is contrasted with character-based BLEU, providing new insight into translation quality assessment.

\item The ``Paradox of Poetic Intent'' is formally introduced as a contradiction between literal fidelity and poetic nuance. This concept illuminates limitations in existing CNLP approaches, particularly in the processing of literary and culturally embedded texts such as Hua Yao.
\end{itemize}

In sum, this work probes the semantic bottlenecks in Chinese–English translation, offering an empirical benchmark and conceptual framework to guide future CNLP development. The findings offer actionable insights for enhancing both bidirectional translation quality and culturally sensitive language modeling.

\section{Corpus Construction and Linguistic Characteristics}
\label{corpus}

This section presents the core corpora used in this study, detailing its sources, textual genres and linguistic features. The dataset encompasses historical dilemmas in scientific terminology translation, modern academic abstracts and metaphorically rich literary lyrics. Together, these form a multi-level, cross-domain Chinese corpus used to evaluate back-translation performance in large language models.

\subsection{Corpus of the ``Xue Dejiong’s Dilemma''}

Historically, scholars have debated whether the Chinese naming of heterocyclic organic compounds should be semantic-based or phonetic-based. He Juan \cite{hejuan2019nomencl} conducted a retrospective review of this issue. Building on that, the present study compiles statements by chemist Xue Dejiong into what is now referred to as the ``Xue Dejiong’s Dilemma'' corpus \cite{weigang2025xdj}. The original Chinese passage is provided in Table \ref{tab:xue_dejiong} and the English translation below was produced by ChatGPT 4.5:

\textit{Xue De-jiong had been dissatisfied with the use of abbreviated heterocyclic compound names containing the Chinese radical \begin{CJK*}{UTF8}{gbsn}``口''\end{CJK*} (mouth radical) for "over ten years," criticizing them as being "as inexplicable as the Buddhist mantra 'Om Mani Padme Hum'." He argued that if phonetic transliterations with mouth radicals were employed for naming compounds such as dithiadiazole, it would inevitably result in repetitive and nonsensical sequences like \begin{CJK*}{UTF8}{gbsn}``\textbf{噁噁噻噻, 唑唑嗪嗪}''\end{CJK*}, making the entire text sound bizarre, like "a foreign scripture." Even a literary genius like Li Bai would find them difficult to comprehend. Thus, he concluded that such simplified phonetic names containing the mouth radical must be abandoned ("painful though it may be"). However, despite his profound dislike for these mouth-radical-based names, he acknowledged: "Many fused heterocyclic structures are exceedingly complex, causing their systematic names to be excessively lengthy. For convenience, certain simplified names are indeed necessary." Yet, he recommended that when creating simplified phonetic characters derived from original pronunciations, at least one character clearly indicating the primary functional group should be appended at the end. For instance, caffeine would appropriately be named \begin{CJK*}{UTF8}{gbsn}``\textbf{咖啡碱}''\end{CJK*} (with \begin{CJK*}{UTF8}{gbsn}``碱''\end{CJK*} indicating its alkaloid nature).}

This passage interweaves technical, cultural and rhetorical elements, reflecting the deep divide among chemists on whether to prioritize literal meaning or phonetic transcription in chemical nomenclature. It thus serves as an ideal benchmark for evaluating machine translation performance \cite{hejuan2019nomencl,weigang2025xdj}.

\subsection{CNKI Abstract Corpus in Scientific Literature}
Scientific publication data were collected from the international portal of the China National Knowledge Infrastructure (oversea.cnki.net), including journal names, article titles, authors, keywords, abstracts, volume and page numbers. A domain-general corpus was then constructed based on ten scientific and technological themes: chemistry, biotechnology, nanotechnology, telemedicine, artificial intelligence, data science, digital economy, linguistics, sociology and distance education.

For each domain, 295 Chinese-language scientific papers were selected and their titles, keywords and abstracts were compiled. All data have been published on GitHub for future research use. The ten disciplines were loosely ordered according to increasing difficulty in language processing. Each subcorpus consists of 295 samples, ensuring statistical validity in subsequent analyses.
Due to space constraints, this study focuses on a chemistry-specific subcorpus comprising 295 abstracts, from which 89 entries were randomly selected to form the \textit{CNKI-CHE-89} corpus. One example, labeled \textit{CNKI-CHE-89-18} in Chinese \cite{feng2024cjac}, is listed in Table \ref{tab:cnki_che_89_18}. The English translation below was generated by Sogou Translate:

\textit{The field of chemical engineering involves a wide range, the whole process is complex and there are many calculation links involved. Calculations in the field of chemical engineering are mostly related to curve fitting nonlinear equations, partial differential, linear or nonlinear programming, etc. These calculations are extremely difficult and it is difficult to achieve accurate and fast calculations only by hand. Combined with computer-aided programming, carrying out metrological calculation can improve the quality and efficiency of applied chemical calculation dose, which has high application value and practical significance.}

\subsection{Corpus from Dao Lang’s Hua Yao Lyrics}

Dao Lang is a Chinese pop singer and musician known for incorporating folklore and literary aesthetics into his work. His song \textit{Hua Yao} (\begin{CJK*}{UTF8}{gbsn}``花妖''\end{CJK*}, "Flower Fairy") centers on the theme of human-demon love tribulation and the timeless pursuit of true love. The lyrics function as a lyrical epic, interlacing poetic sentiment, historical geography, Buddhist cosmology and nonlinear temporality. They depend heavily on culturally unique semiotic systems, such as \textit{I Ching} directional philosophy and Taoist geomantic texts like the \textit{Compass Classic} (\begin{CJK*}{UTF8}{gbsn}``\textbf{罗盘经}''\end{CJK*}); on narrative logic involving temporal recursion (past life and present life); and on prosodic aesthetics, including tonal counterpoint and phonetic ambiguity. These features present triple challenges for machine translation:

\begin{itemize}
\item	Symbolic deconstruction of cultural metaphors (e.g., reducing ``Compass Classic'' to a mere navigational tool);
\item Detachment from historical spatiotemporal context (e.g., translating place names into empty referents);
\item Loss of poetic rhythm and structural elegance, with rhymed parallel structures being flattened into prosaic narration. 
\end{itemize}

These translation challenges highlight structural limitations in current NLP models when dealing with cross-cultural interpretation and literary transcreation. For this reason, Hua Yao was chosen as the centerpiece of the study’s literary back-translation task. The Chinese original is provided in Table \ref{tab:hua_yao} and the following English version was produced using Google Translate:

\textit{Dao Lang, part of the lyrics of "Flower Demon": "I am the wandering tears on the annual rings, you can still smell the rouge in the wind, if I carve my promise on the river bank, the river is cold and the moonlight fills the city, I have waited for you for a long time under the tree of time, the mortal world entangles me, slanders me, laughs at me and my hair turns white, you see the kite chasing the sunset on the horizon, like a wind lantern looking back to say goodbye in the middle of the night, my heart is like quicksand exiled beside the wheel rut, if you come back one day, you must be wandering in the world, if you encounter that autumn night rain, the birds are also wet, but it is the withered yellow lingering under the flower wall, you live in the east of Qiantang, I live in the north of Lin'an, you left with brown clothes and red, my little slave's waist is yellow, I misplaced the compass and was thrown into the wrong place at Quanting, I went to Hangzhou and you were born in Yuhang again."}

The corpora introduced in this section addresses linguistic complexity and domain diversity through three primary text types: historical texts reflecting terminological controversies, scientific abstracts from CNKI and culturally dense literary lyrics. These corpora highlight distinct challenges in machine translation, such as semantic compression, cultural decoding and syntactic fidelity. Together, they provide a contrastive and structurally complementary framework for evaluating back-translation performance in subsequent experiments.

\section{Methodology}

This section introduces the back-translation method, systematically elaborating its application in Chinese NLP in terms of basic concepts, overall framework, implementation steps, language models and tools and evaluation metrics.

\subsection{ Overall Framework of Back-Translation}
Back-Translation, also known as reverse translation, is a method for objectively evaluating translation accuracy \cite{tu2017review,artetxe2018unsupervised, ozolins2020translation}. It is a significant research method in both machine translation (MT) and natural language processing (NLP), especially suited for data augmentation, translation quality assessment and cross-lingual semantic alignment in complex language systems such as Chinese and Japanese. Its core idea is to translate the source language (e.g., Chinese, ZH) into a target language (e.g., English, EN) and then translate it back to the source language. By comparing the original and back-translated texts, the performance of translation systems or language models can be analyzed and optimized.

The process of implementing Chinese back-translation (ZHx → EN → ZHy) can be divided into three stages, also see Figure~\ref{imagery_vs_selfawareness}:
\begin{itemize}
   
 \item Forward Translation (ZHx → EN): Translate the original Chinese text using a translation system (e.g., NMT models or LLMs) to produce an English version.

 \item Back Translation (EN → ZHy): Translate the English text back into Chinese, yielding the back-translated text.

 \item Comparative Analysis: Compare Zhx and ZHy using automatic metrics (e.g., BLEU, TER) and human evaluation (e.g., semantic consistency, fluency, cultural adaptation) to quantify deviations between the original and back-translated versions.

\end{itemize}

The theoretical foundation of this method lies in the Translation Equivalence Theory \cite{nida1964toward} and the Cyclic Consistency Hypothesis \cite{artetxe2018unsupervised}, which posit that high-quality translation systems should maintain semantic integrity between source and target texts. If the back-translated version significantly deviates from the original (e.g., BLEU $\leq 0.30$ or human score $ \leq 3/5$), this suggests information loss or cultural distortion. 

With the advancement of MT and AI technologies, back-translation has expanded into the following research areas:
\begin{itemize}
    
\item Evaluation of Translation Tools: Use ZHx → EN → ZHy loop tests to compare the Chinese processing capabilities of systems like Google Translate and mainstream LLM platforms.

\item Analysis of Translation Strategies for Technical Terms: Quantify the retention of transliteration (e.g., ``\begin{CJK*}{UTF8}{gbsn}比特\end{CJK*}'' from ``bit") vs. semantic translation (e.g., ``\begin{CJK*}{UTF8}{gbsn}人工智能\end{CJK*}'' for ``artificial intelligence") across languages.

\item Semantic Representation of Chinese Characters and Words: Analyze vector-space mappings through embedding similarity (e.g., cosine similarity), including traditional input methods such as Cangjie and Wubi, as well as recent Six-Writings phonosemantic encodings introduced into NLP for enhanced language processing \cite{weigang2024six}.
 
\end{itemize}

In back-translation, some MT models, especially large pre-trained LLMs (e.g., DeepSeek V3 \cite{liu2024deepseek}, ChatGPT 4.5), tend to output Chinese texts that closely mirror the original rather than producing independently translated texts. This phenomenon is termed mirrored back-translation. For instance, in experiments using the lyrics of Hua Yao, DeepSeek V3's output nearly replicated the original (BLEU 0.7602), suggesting that the model might be bypassing actual translation via memorization. This mirrored behavior both challenges traditional evaluation and hints at the ``Quasi-self-awareness" potential of LLMs.

The \textit{Paradox of Poetic Intent} is a new concept proposed in this paper. It refers to LLMs’ tendency to prioritize literal consistency at the expense of deeper semantic and cultural meaning in Chinese texts. Chinese mainly uses one- and two-character words (accounting for ~95\% of the lexicon \cite{liu1986chinese}), but three- and four-character expressions (e.g., idioms, set phrases), though less frequent, carry significant meaning and imagery. For example, in \textit{Hua Yao}, the phrase \begin{CJK*}{UTF8}{gbsn}``夤夜的风灯''\end{CJK*} ("wind lanterns of the late night") conveys a poignant farewell image, which LLMs often flatten into "midnight lantern," losing its emotional depth. Most LLMs rely on character-based preprocessing, limiting their ability to capture phrasal semantic relations and exacerbating the \textit{Imagery Paradox} in Chinese translation.

\subsection{Implementation Steps of Back-Translation}

This study evaluates the performance of large language models (LLMs) on Chinese back-translation tasks, focusing on semantic preservation, terminological consistency and cultural expression. The workflow is divided into three stages: data preparation, model selection and evaluation metrics.

\subsubsection{Data Preparation}
To ensure coverage of diverse text types and linguistic styles, we use three categories of source materials:
\begin{itemize}
\item Literary Texts: Representative works like the lyrics of Flower Demon by Dao Lang are selected to test the models’ handling of metaphor, rhetoric and imagery.

\item Scientific Abstracts: A sample of 295 Chinese abstracts from CNKI (covering science and engineering) is used to assess terminological alignment and logical consistency.

\item Translation Paradox Cases: Texts known for semantic asymmetry in C-E translation (e.g., the ``Xue Dejiong’s dilemma") are used to test model sensitivity to cross-lingual challenges.
 
\end{itemize}

\subsubsection{ Model Selection}
Two categories of models are compared:
\begin{itemize}
   
 \item Neural Machine Translation (NMT): Google Translate, Baidu Translate and Sogou Translate, representing current industrial standards.

 \item Large Language Models (LLMs): Six leading platforms, GPT-4.5, Claude 3.7, DeepSeek V3, Gemini 2.0, Grok 3 and Moonshot, are tested to compare their bidirectional performance in ZH-EN translation.
 
\end{itemize}

All models use open APIs or web interfaces. The unified pipeline is ZHx→EN→ZHy, with special attention to mirrored back-translation behavior.

 \subsubsection{Multi-Dimensional Evaluation}
Evaluation covers four key dimensions (see Table \ref{tab: methodology}):

\begin{table*}[htbp]
\centering
\caption{Evaluation Dimensions of Chinese Back-Translation Quality}
\begin{tabular}{cp{5cm}p{5cm}}
\hline
\textbf{Dimension} & \textbf{Evaluation Focus} & \textbf{Key Metrics} \\
\hline
Semantic Consistency & Alignment of core meaning between original and back-translated texts & Semantic Similarity (SS) \\

Fluency & Grammatical correctness and natural expression in back-translated texts & TER (Translation Edit Rate) \\

Cultural Fidelity & Accurate transmission of culture-laden words (e.g., idioms, imagery) & Human expert evaluation (e.g., literary translators) \\

Terminological Consistency & Accurate alignment of scientific and technical terms between Chinese and English & BLEU, CHRF and term-alignment checks \\
\hline
\end{tabular}
\label{tab: methodology}
\end{table*}

To enhance objectivity, both single-sample qualitative and multi-sample statistical analyses are conducted. The Friedman test \cite{demvsar2006statistical} is used to identify significant differences across models, with visualizations (e.g., heatmaps) supporting the interpretation.

\subsection{Assessing BLEU Score Variability in LLMs - A Non-Parametric Comparison Using the Friedman Test}

This subsection outlines the method for evaluating LLMs and translation tools using back-translation, with the BLEU score as the quality metric.

Each text selected for the experiment is considered an independent block, with \( n \) representing the number of texts (blocks). Each text is independently processed by all \( k = 5 \) state-of-art models (Grok-beta, DeepSeek-V3\footnote{DeepSeek-R1 does not translate ZH$\rightarrow$EN, presenting reasoning only in ZH. Therefore, DeepSeek-V3 was chosen.}, GPT 4.5, Gemini 2.0 and Claude 3.7). Texts must reflect diverse content to capture real variability in translation scenarios.

In the experimental procedure, each model translates the text from the source language to an intermediate language and then back to the original language. The BLEU score of the back-translation serves as a measure of semantic and syntactic preservation. The BLEU scores, represented by \( Y_{ij} \) for the \( j \)-th text (block) translated by the \( i \)-th model, are organized into a data matrix with \( k = 5 \) treatments (models) and \( n \) blocks (texts). To evaluate translation consistency, each text undergoes \( r = 3 \) translation repetitions per model, conducted in fully independent sampling sessions using randomized seeds and prompt variants, thereby ensuring statistical independence across runs and allowing robust estimation of intra-model variability.

The analysis aims to distinguish translation model effects while accounting for text variability, with the model formulated as:
\begin{equation}
    Y_{ij} = \eta + \tau_i + \beta_j + \epsilon_{ij},
\end{equation}
where:
\( \eta \) represents the overall mean BLEU score, \( \tau_i \) denotes the effect of the \( i \)-th model, \( \beta_j \) accounts for the effect of the \( j \)-th text, controlling for inherent text variability and \( \epsilon_{ij} \) corresponds to the random error associated with the translation of text \( j \) by model \( i \).

Due to potential violations of normality and homoscedasticity, non-parametric methods are employed. Specifically, the Friedman test is applied to detect significant differences between translation models across texts. When significant differences are observed, the Dunn post-hoc test with adjustments for multiple comparisons (e.g., Bonferroni correction) is used to identify statistically distinct model pairs \cite{Sheldon1996, Aqmal2024, Arruda2020, Nam2015}.

The sample size was calculated for a significance level of 0.05 and a power of 0.8, considering a medium effect size (\( f = 0.3 \)) for a repeated-measures ANOVA model. This method was chosen due to the absence of widely accepted power analysis procedures for the Friedman test. It required \( n = 89 \) distinct texts, each translated by \( k = 5 \) models, with \( r = 3 \) repetitions per text, totaling 1,335 observations. All texts used in this evaluation correspond to scientific abstracts from the field of chemistry, ensuring genre homogeneity and minimizing cross-domain variance. During the preliminary statistical checks, empirical violations of normality and homoscedasticity, assessed via the Shapiro–Wilk and Levene’s tests, led to the adoption of the Friedman test for robustness in the subsequent analysis.

The selected metrics capture different aspects of machine translation quality and were chosen to provide a comprehensive evaluation of lexical accuracy, structural integrity and semantic preservation. Each metric addresses specific limitations of traditional evaluation methods, ensuring a balanced assessment of translation performance.

Bilingual Evaluation Understudy (BLEU\footnote{We apply two BLEU configurations: the standard BLEU with weights $(0.5, 0.5, 0.0, 0.0)$ and the uniform-weight BLEU-Unif with $(0.25, 0.25, 0.25, 0.25)$ over 1 to 4-grams. The rationale is to evaluate technical terms more precisely using BLEU, while BLEU-Unif captures broader patterns, such as those introduced through back-translation.}
) \cite{papineni2002bleu}, Character F-score (CHRF) and Semantic Similarity (SS) follow a higher is better interpretation, while Translation Edit Rate (TER) follows a lower is better approach. BLEU measures n-gram overlap, with higher scores indicating greater lexical similarity, but struggles with paraphrasing, where meaning is retained but wording differs. CHRF evaluates character-level similarity, making it effective for morphologically rich languages like Chinese. SS assesses meaning preservation using Term Frequency-Inverse Document Frequency (TF-IDF), capturing relationships missed by BLEU and CHRF. TER quantifies the editing effort (insertions, deletions and substitutions) needed to match the reference, where lower scores indicate better performance.

These metrics were selected to balance structural fidelity and semantic accuracy. BLEU and CHRF provide insight into lexical and character-level similarity, while TER captures translation fluency by measuring the required modifications. SS complements these by evaluating semantic retention, mitigating the limitations of token-based metrics.

For Chinese, character segmentation initially led to errors with compound words, reducing metric reliability. The adoption of Jieba, a specialized Chinese tokenizer, improved tokenization accuracy, ensuring BLEU, CHRF, TER and SS operate on coherent linguistic units, enhancing evaluation consistency.

\begin{table*}[htbp]
\centering
\caption{Comparison of Back-Translation Quality Across Different Models and Corpora}
\begin{tabular}{lcccccc}
\toprule
\multirow{2}{*}{LLM/Tool} & \multicolumn{2}{c}{XDJ's Dilemma} & \multicolumn{3}{c}{CNKI-CHE-89-18} & DaoLang \\
\cmidrule(lr){2-3} \cmidrule(lr){4-6} \cmidrule(lr){7-7}
 & GrokBleu & GrokBleu & GPTBleu & JebaBleu & SD & JebaBleu \\
\midrule
Grok xAI & 0.65 & 0.61 & 0.34 & 0.2786 & 0.0583 & 0.3311 \\
DeepSeek V3 & 0.64 & 0.58 & 0.37 & 0.3042 & 0.0576 & 0.7602 \\
ChatGPT 4.5 & 0.59 & 0.59 & 0.36 & 0.2970 & 0.0000 & 0.2141 \\
ChatGPT 4 & 0.47 & 0.50 & - & - & - & - \\
Gemini 2.0 & 0.51 & 0.52 & 0.40 & 0.3382 & 0.0016 & 0.2506 \\
Claude 3.7 & - & - & 0.42 & 0.3702 & 0.0099 & 0.3114 \\
Mistral & 0.24 & 0.46 & - & - & - & - \\
Google Trans. & 0.33 & 0.44 & 0.45 & 0.4719 & - & 0.3664 \\
Baidu Translate & 0.25 & 0.36 & 0.43 & 0.3769 & - & 0.2530 \\
Sougo Translate & 0.17 & 0.31 & 0.57 & 0.5212 & - & 0.3650 \\
\bottomrule
\end{tabular}
\label{tab:backtranslation-bleu}
\end{table*}

\section{Comparison of Back-Translation Quality and Model Performance Across Platforms}
This study evaluates the back-translation capabilities of nine mainstream LLMs and translation tools using three representative Chinese text samples: Xue Dejiong’s Dilemma, CNKI-CHE-89-18 and lyrics from Dao Lang’s Hua Yao. Two categories of models are assessed: (1) traditional translation tools (Google, Baidu, Sogou) and (2) large language models (ChatGPT, Grok, DeepSeek, Gemini, Claude and Mistral). All systems are publicly available and allow reproducible testing.

\subsection{Overall Model Performance}
All samples were translated from Chinese to English and then back to Chinese. BLEU scores were calculated to assess fidelity. Table \ref{tab:backtranslation-bleu} shows back-translation scores across platforms, where BLEU scores were computed on Grok 3 (GrokBleu), GPT-4 (GPTBleu) and a Jieba-based BT-Fried\footnote{\url{https://nlpmetriclab.streamlit.app/}} system (JebaBleu). Xue Dejiong’s Dilemma contains complex academic, cultural and rhetorical elements. CNKI-CHE-89-18 is a simplified scientific abstract. Hua Yao corpus includes poetic lyrics.

For Xue Dejiong’s Dilemma, Grok xAI and DeepSeek V3 achieved the highest BLEU scores (0.65 and 0.64). A BLEU score of $\geq 0.60$ in literary texts is notably high compared to the typical range of 0.30–0.50 for such tasks, indicating strong surface-level fidelity but potential loss of poetic nuance. The high BLEU score may stem from DeepSeek-V3’s tendency to prioritize surface-level matching over semantic transformation, possibly due to its training data containing similar scientific texts. In contrast, traditional tools performed worse, especially on academic texts (e.g., Baidu and Sogou scored $\leq 0.30$).  

In the CNKI-CHE-89-18 test, Sogou achieved the best BLEU score (0.57) under GPT4-based evaluation. Google and Baidu scored 0.45 and 0.43, respectively. Meanwhile, Grok, DeepSeek and GPT-4.5 scored lower than 0.40. Claude 3.7 was the best among LLMs (0.42).

A JebaBleu metric, computed using Jieba segmentation, was introduced as a reference to LLM-based BLEU scores. Although values varied slightly, ranking consistency was largely maintained. Table \ref{tab:backtranslation-bleu} shows the back-translation BLEU scores across platforms (ZHx $\rightarrow$ EN $\rightarrow$ ZHy).

\subsection{Case Study: Xue Dejiong’s Dilemma} 

This chemically-themed text features technical terms (e.g., \begin{CJK*}{UTF8}{gbsn}``噁噁噻噻''\end{CJK*}), Buddhist mantras, cultural allusions (e.g., \begin{CJK*}{UTF8}{gbsn}``谪仙李太白''\end{CJK*}) and emotional expressions, making it challenging even for native Chinese speakers \cite{hejuan2019nomencl}.

Table~\ref{tab:xue_dejiong} shows the comparison of Back-Translation examples for Xue Dejiong's Dilemma, especially for the sentence: \textbf{\begin{CJK*}{UTF8}{gbsn}满纸咿哑，一若番书，虽有谪仙李太白其人，恐亦难于索解。\end{CJK*}} There are the English (EN) and Chinese (ZHy) translations by every LLM. The results show that Grok xAI better retained semantic and cultural elements during back-translation, whereas Mistral and Google Translate produced inconsistent or inaccurate renderings. Sogou’s translation showed semantic gaps and logic flaws, resulting in a BLEU score of only 0.17.

\begin{table*}[ht]
\small
\centering
\caption{Comparative Analysis of Back-Translation Instances for the ``Xue Dejiong Dilemma'' across Different Language Models}
\label{tab:xue_dejiong}
\begin{tabular}{lp{4cm}p{3.8cm}p{3.8cm}c}
\toprule
Model & Chinese Original & English Translation & Back-Translated Chinese & BLEU \\
\midrule
Grok xAI & \multirow{8}*{\parbox{4.2cm}{\begin{CJK*}{UTF8}{gbsn}薛德炯 ``不满于'' 口旁简名 ``已十余年'', 认为它们 ``像佛经中的 '唵嘛呢叭咪吽', 简直莫名其妙''！如若用口旁音译字来命名 dithiadiazole 等杂环化合物，势必 ``噁噁噻噻, 唑唑嗪嗪, \textbf{满纸咿哑, 一若番书, 虽有谪仙李太白其人, 恐亦难于索解。}'' 此口旁之简名所以不能不革除（割爱）也。即便如此深恶痛绝口旁名称, 也承认 ``稠杂圜之结构繁复者多, 故其系统名类多冗长。为便于名举计, 确有特定简名之必要''。不过他建议 ``从原名之音, 特创口旁字之简名, 至少须于名末多加一个足以表示其主要官能之字''。如咖啡宜命名为 ``咖啡碱''。\end{CJK*}}} & filling the page with gibberish, as if it were a foreign text. Even someone as extraordinary as the banished immortal Li Taibai (Li Bai) would likely find it difficult to decipher. & \begin{CJK*}{UTF8}{gbsn}满纸都是胡言乱语,仿佛外文书籍。即使有谪仙李太白（李白）这样杰出的人物,恐怕也难以理解。\end{CJK*} & 0.65 \\
DeepSeek V3 & & crowding the page with gibberish, as cryptic as foreign script. Even a genius like Li Bai, the Banished Immortal, would struggle to decipher them. & \begin{CJK*}{UTF8}{gbsn}满纸咿哑,一若番书,纵有谪仙李太白其人,恐亦难于索解。\end{CJK*} & 0.64 \\
GPT 4.5 & & making the entire text sound bizarre, like ``a foreign scripture.'' Even a literary genius like Li Bai would find them difficult to comprehend. & \begin{CJK*}{UTF8}{gbsn}使整个文本听起来十分古怪,``一若番书''。即便像李白这样才华横溢的文学天才,也难以理解。\end{CJK*} & 0.59 \\
GPT 4 & & a chaotic jumble of syllables filling the pages, resembling an unintelligible foreign script. ``Even if the great poet Li Bai himself were present, he would struggle to decipher them.'' & \begin{CJK*}{UTF8}{gbsn}整个页面充满混乱的音节,看起来就像一篇难以辨认的外文。``即便谪仙李白在世,也恐怕难以解读。''\end{CJK*} & 0.47 \\
Google & & it will inevitably be ``full of babbling, like foreign language'' and even the exiled immortal Li Bai would find it difficult to understand. & \begin{CJK*}{UTF8}{gbsn}必然``咿呀学语,犹如外语,连流放的仙人李白都听不懂''\end{CJK*} & 0.33 \\
Baidu & & full of paper babbling, like a book; even if there is an exiled immortal Li Taibai, it may be difficult to understand. & \begin{CJK*}{UTF8}{gbsn}满纸胡言乱语,像一本书,即使有流亡的仙人李太白,也可能很难理解。\end{CJK*} & 0.25 \\
Mistral & & filling the page with strange symbols, like a foreign script. Even the exiled immortal Li Bai would struggle to decipher them. & \begin{CJK*}{UTF8}{gbsn}满纸奇怪的符号,像外文一样。即使是被贬谪的仙人李白也难以解读。\end{CJK*} & 0.24 \\
Sogou & & it will inevitably be ``full of babbling, like foreign language,'' and even the exiled immortal Li Bai would find it difficult to understand. & \begin{CJK*}{UTF8}{gbsn}论文哑了。如果是一本书,即使有一个堕落的仙女李太白,也很难找到。\end{CJK*} & 0.17 \\
\bottomrule
\end{tabular}
\vspace{1mm}
\footnotesize{Note: BLEU values are calculated by the Grok 3 platform.}
\end{table*}

\subsection{Case Study: CNKI-CHE-89-18}

This scientific abstract uses everyday language to express technical achievements. While forward translation was generally handled well, reverse translation struggled with idiomatic expressions like \begin{CJK*}{UTF8}{gbsn}提质增效\end{CJK*} (``Improve quality and efficiency'').

From Table~\ref{tab:cnki_che_89_18}, we find that:
\begin{itemize}
 \item Traditional tools (Google, Baidu, Sogou) outperformed LLMs, thanks to: word-by-word translation and structural fidelity; optimizations for Chinese users; conservative generation strategies. Only Sogou and Baidu translated correctly the term \begin{CJK*}{UTF8}{gbsn}化学计算剂量\end{CJK*} (``Chemical calculation dose'').

 \item LLMs struggled due to: Overgeneralization of Chinese-specific expressions; Semantic drift in technical terms; Inconsistencies from random generation.

 \item Cultural expressions like \begin{CJK*}{UTF8}{gbsn}提质增效\end{CJK*} (``Improve quality and efficiency'') were often rephrased or lost in LLMs, while traditional tools preserved approximate meaning via literal rendering.
\end{itemize}

\begin{table*}[ht]
\centering
\small
\caption{Comparative Analysis of Back-Translation Instances for the ``CNKI-CHE-89-18'' Corpus across Different Language Models}
\label{tab:cnki_che_89_18}
\begin{tabular}{lp{4cm}p{4cm}p{4cm}c}
\toprule
Model & Chinese Original & English Translation & Back-Translated Chinese & BLEU \\
\midrule
Sogou & \multirow{8}*{\parbox{4cm}{[\begin{CJK*}{UTF8}{gbsn}正\end{CJK*}] \begin{CJK*}{UTF8}{gbsn}化学工程领域的涉及面较广，全过程复杂，涉及到的计算环节众多。化学工程领域的计算多与曲线拟合、非线性方程、偏微分、线性或非线性规划的求解等有关，这些计算难度极大，仅靠手工难以实现精准快捷的计算；而配合计算机辅助编程，开展计量计算，能够实践应用化学计算剂量的提质增效，具有极高的应用价值和现实意义。\end{CJK*}}} 
& carrying out metrological calculation can improve the quality and efficiency of applied chemical calculation dose. 
& \begin{CJK*}{UTF8}{gbsn}进行计量计算可以提高应用化学计算剂量的质量和效率。\end{CJK*} & 0.57 \\
Google & & the measurement calculation can be carried out to improve the quality and efficiency of chemical calculation dosage in practice. 
& \begin{CJK*}{UTF8}{gbsn}可以大大提高化学计算的质量和效率。\end{CJK*} & 0.45 \\
Baidu & & conducting quantitative calculations can improve the quality and efficiency of chemical dosage calculations. 
& \begin{CJK*}{UTF8}{gbsn}进行定量计算可以提高化学剂量计算的质量和效率。\end{CJK*} & 0.43 \\
Claude & & the quality and efficiency of chemical dosage calculations can be improved in practical applications. 
& \begin{CJK*}{UTF8}{gbsn}可以在实际应用中提高化学计量计算的质量和效率。\end{CJK*} & 0.42 \\
Gemini 2.0 & & performing quantitative calculations can practically improve the quality and efficiency of chemical dosage calculations. 
& \begin{CJK*}{UTF8}{gbsn}进行定量计算能够有效提高化学计量计算的质量和效率。\end{CJK*} & 0.40 \\
DeepSeek & & it is possible to enhance the quality and efficiency of chemical dosage calculations in practical applications. 
& \begin{CJK*}{UTF8}{gbsn}能够在实际应用中提升化工配比计算的质量与效率。\end{CJK*} & 0.37 \\
GPT 4.5 & & it becomes possible to improve the quality and efficiency of chemical dosage calculations in practical applications. 
& \begin{CJK*}{UTF8}{gbsn}可以提高实际应用中化学加药计算的质量和效率。\end{CJK*} & 0.36 \\
Grok xAI & & it is possible to enhance the efficiency and effectiveness of chemical dosage calculations. 
& \begin{CJK*}{UTF8}{gbsn}可以提高化学配比计算的效率和效果。\end{CJK*} & 0.34 \\
\bottomrule
\end{tabular}
\vspace{1mm}
\footnotesize{Note: BLEU values are calculated by the Grok 3 platform.}
\end{table*}

\subsection{Case Study: Hua Yao Lyrics}

A key line from the lyrics of \textit{Hua Yao} reads: ``When you left, your coarse robe turned red; the yellow sash adorned my waist. The Compass Classic was misread and I was wrongly cast into Quanting'' 
(\begin{CJK*}{UTF8}{gbsn}君去时褐衣红，小奴家腰上黄，寻差了罗盘经，错投在泉亭\end{CJK*}). 
We note this sentence as the ``Hua Yao Red and Yellow'' corpus 
(\begin{CJK*}{UTF8}{gbsn}“花妖红黄”语料\end{CJK*}).

The term ``coarse robe'' (\begin{CJK*}{UTF8}{gbsn}褐衣\end{CJK*}) refers to garments made of rough cloth, indicating that the young man came from a humble background. The ``yellow sash'' was an ornamental accessory worn around the waist by noblewomen during the Song Dynasty, typically belonging to women from wealthy families. Against this backdrop, the noblewoman’s father could not tolerate the relationship and killed the impoverished lover with a blade, hence the line ``your coarse robe turned red'' (\begin{CJK*}{UTF8}{gbsn}褐衣红\end{CJK*}), vividly portraying the moment his blood stained his clothing. The following line, ``the yellow sash adorned my waist'' (\begin{CJK*}{UTF8}{gbsn}腰上黄\end{CJK*}), further implies that the woman used this same sash to take her own life in devotion to her beloved.

This section uses this particular lyric as a case study to examine the capability of large language models (LLMs) to process texts that are rich in cultural allusions, historical references, and poetic resonance.

\subsubsection{Metric-Based Evaluation}

Table \ref{tab:flower-backtranslation} shows five evaluation metrics (BLEU, BLEU-Unif, CHRF, TER, Semantic Similarity) for Hua Yao by BT-Fried system. DeepSeek-V3 achieved outstanding results (BLEU 0.7602), significantly outperforming all others, suggesting verbatim back-translation, where the model recalls or reconstructs the original instead of translating.

To validate this, the same EN output by DeepSeek-V3 was manually back-translated using DeepSeek-V3 via human prompting, yielding a BLEU of just 0.2849. This contrast with the automatic BLEU score of 0.7602 suggests that the original model output was not a result of semantic reconstruction, but rather of memorization or internal matching. While BLEU scores capture surface-level similarity, they are insufficient to confirm deeper semantic or stylistic fidelity. Therefore, interpretations related to quasi-verbatim translation or the Paradox of Poetic Intent should be understood in conjunction with qualitative assessments and comparative analysis across model behaviors.

\begin{table}[h!]
\centering
\caption{Back-Translation Evaluation Metrics on \textit{Hua Yao} Lyrics}
\begin{tabular}{lccccc}
\hline
\textbf{Model} & \textbf{BLEU} & \textbf{BLEU-Unif} & \textbf{CHRF} & \textbf{TER} & \textbf{SS} \\
\hline
DeepSeek-V3    & 0.7602 & 0.6481 & 0.9184 & 0.0253 & 0.6457 \\
DeepThink R1   & 0.6096 & 0.4893 & 0.8571 & 0.0378 & 0.3558 \\
Google Trans.   & 0.3664 & 0.1347 & 0.7551 & 0.0400 & 0.0423 \\
Sogou Trans.    & 0.3650 & 0.2063 & 0.5918 & 0.0434 & 0.0460 \\
Grok-beta       & 0.3311 & 0.1501 & 0.6531 & 0.0392 & 0.0203 \\
Claude-3.7      & 0.3114 & 0.1449 & 0.6463 & 0.0398 & 0.0648 \\
DeepSeek-V3\#   & 0.2849 & 0.1409 & 0.6259 & 0.0423 & 0.0690 \\
Baidu Trans.    & 0.2530 & 0.1045 & 0.6939 & 0.0384 & 0.0200 \\
Gemini-2.0      & 0.2506 & 0.0545 & 0.6463 & 0.0392 & 0.0000 \\
GPT-4.5         & 0.2141 & 0.0523 & 0.5306 & 0.0434 & 0.0000 \\
\hline
\end{tabular}
\label{tab:flower-backtranslation}
\\
\footnotesize{Note: BLEU used (0.5, 0.5, 0, 0) weighting; BLEU-Unif used uniform (0.25, 0.25, 0.25, 0.25). SS is the Semantic Similarity.}
\end{table}

\subsubsection{Translation Instance Comparison} 

Among the most challenging elements in Chinese–English literary translation are phrases rooted in traditional cosmology, religious symbolism, or regionally embedded beliefs. A notable example appears in the last sentence of the ``Hua Yao Red and Yellow'' corpus (\begin{CJK*}{UTF8}{gbsn}“花妖红黄”语料\end{CJK*}): ``The Compass Classic was misread and I was wrongly cast into Quanting.'' (\begin{CJK*}{UTF8}{gbsn}寻差了罗盘经，错投在泉亭\end{CJK*})

At first glance, this line might be interpreted as a simple navigational error. However, a deeper reading reveals its embedded cultural logic. In Taoist and folk belief systems, the \textit{Compass Classic} (\begin{CJK*}{UTF8}{gbsn}罗盘经\end{CJK*}) is not a literal compass, but a geomantic scripture that encodes metaphysical directions and reincarnation timing. The phrase ``misreading the Compass Classic'' implies that a netherworld agent (often referred to in Chinese folklore as a ``little ghost'' or \begin{CJK*}{UTF8}{gbsn}阴差\end{CJK*}) mishandled the timing of the girl’s reincarnation.

The word \textit{Quanting} (\begin{CJK*}{UTF8}{gbsn}泉亭\end{CJK*}), although literally translatable as ``Spring Pavilion,'' holds a far more specific historical meaning: it was the name for Hangzhou during the Western Han dynasty. In contrast, \textit{Yuhang}, the reincarnation site of the male lover in the lyrics, corresponds to another historical name for the same region in a different era. Thus, the girl and the boy are both reincarnated in what is geographically the same city, but in entirely different historical periods.

This ``mismatched casting'' conveys a poignant metaphor: though their souls return to the same earthly location, they are forever separated by time, fated never to reunite. It reflects a deeper temporal tragedy encoded through historical toponyms, a nuance easily missed by machine translation systems.

In our April 2025 back-translation experiments, all major LLMs (including GPT-4.5, Claude 3.7, Grock 3 and Gemini 2.0) failed to recognize this cultural and historical subtext. Most rendered ``Compass Classic'' as a mere navigation manual and translated ``Quanting'' as an ordinary place name, thereby losing the profound implications of time-displaced fate. DeepSeek-V3 produced a near-verbatim return: BLEU 0.7602. Manual back-translation of the same EN result on DeepSeek-V3\# showed BLEU 0.2849, confirming the memorization-based artifact, see Table \ref{tab:hua_yao}.

This example highlights not only the semantic blind spots of LLMs in processing culturally loaded texts but also emphasizes the importance of temporal semantics and historical reasoning in CNLP systems.

\begin{table*}[ht]
\centering
\small
\caption{Comparative analysis of back-translated versions of Daolang’s ``Hua Yao'' lyrics across various language models}\label{tab:hua_yao}
\begin{tabular}{lp{4cm}p{3.8cm}p{3.8cm}c}
\toprule
Model & Chinese Original & English Translation & Back-Translated Chinese & BLEU \\
\midrule
DeepSeek & \multirow{8}*{\parbox{4cm}{\begin{CJK*}{UTF8}{gbsn}刀郎，《花妖》部分歌词：“我是那年轮上流浪的眼泪，你仍然能闻到风中的胭脂味，我若是将诺言刻在那江畔上，一江水冷月光满城的汪洋，我在时间的树下等了你很久，尘凡儿缠我谤我笑我白了头，你看那天边追逐落日的纸鸢，像一盏回首道别夤夜的风灯，我的心似流沙放逐在车辙旁，他日你若再返必颠沛在世上，若遇那秋夜雨倦鸟也淋淋，那却是花墙下弥留的枯黄，君住在钱塘东，妾在临安北，君去时褐衣红，小奴家腰上黄，寻差了罗盘经，错投在泉亭，奴辗转到杭城，君又生余杭。”\end{CJK*}}} & You left in robes of brown and red, my sash was yellow. Misreading the compass, I strayed to Quan’ting. & \begin{CJK*}{UTF8}{gbsn}君去时褐衣红，我腰上黄。寻差了罗盘经，错投在泉亭。\end{CJK*} & 0.7602 \\
Google & & you left with brown clothes and red, my little slave's waist is yellow, I misplaced the compass and was thrown into the wrong place at Quanting. & \begin{CJK*}{UTF8}{gbsn}你留下褐衣赤色，我的小奴腰间黄，我放错了地方罗盘在泉亭丢错了地方。\end{CJK*} & 0.3664 \\
Sogou & & When you go, your clothes are brown and red and your little slave's waist is yellow. You lost the compass and threw it in the spring pavilion by mistake. & \begin{CJK*}{UTF8}{gbsn}你去的时候，衣服是棕红色的，小奴才的腰是黄色的。你把指南针弄丢了，不小心扔进了春亭。\end{CJK*} & 0.3650 \\
Grok xAI & & When you left, you wore a brown coat, with a yellow belt around my waist, I've lost my way with the compass, mistakenly went to Quan Pavilion. & \begin{CJK*}{UTF8}{gbsn}你离去时身披褐色外套，腰间系黄带，我迷失了指南针，误入泉亭。\end{CJK*} & 0.3311 \\
Claude 3.7 & & When you left you wore a brown and red robe and this little servant had yellow at her waist. I followed the wrong compass, mistakenly arriving at Spring Pavilion. & \begin{CJK*}{UTF8}{gbsn}你离去时穿着褐红袍，这小婢腰间束着黄。我循错了罗盘，误到春台上。\end{CJK*} & 0.3114 \\
Baidu & & when you went, your clothes were brown and red and my waist was yellow. I missed the compass and accidentally threw it at Quanting. & \begin{CJK*}{UTF8}{gbsn}你去的时候，你的衣服是棕红色的，我的腰是黄色的。我错过了指南针，不小心把它扔到了匡亭。\end{CJK*} & 0.2530 \\
Gemini 2.0 & & When you left, your brown clothes were red, the little slave girl's waist is yellow. Searched wrongly for the compass scripture, wrongly cast in Quanting. & \begin{CJK*}{UTF8}{gbsn}你离去时，棕色的衣衫已泛红，小奴婢的腰带是黄的。误寻罗经忏，错铸在泉亭。\end{CJK*} & 0.2506 \\
GPT 4.5 & & When you departed, your robe was red, my waistband yellow; Mistakenly following the compass needle, arriving wrongly at Quanjing. & \begin{CJK*}{UTF8}{gbsn}初识你时你红衣垂落，我腰系黄带；却误随指针错抵权境。\end{CJK*} & 0.2141 \\
\bottomrule
\end{tabular}
\vspace{1mm}
\footnotesize{Note: BLEU scores were computed using the BT-Fried evaluation system.}
\end{table*}

\subsubsection{Discussion and Implications}

Hua Yao features rich temporal, spatial and emotional dimensions. Most models, except DeepSeek-V3 and DeepThink R1, failed to capture these in translation. Literal translations (e.g., ``little slave'') introduced semantic noise. Back-translations were even less faithful.

Interestingly, DeepSeek-V3's ``verbatim back-translation" bypassed actual translation through likely memorization. While it challenges conventional evaluation, it also reveals a form of data-driven awareness. This aligns with the ``Paradox of Poetic Intent'' discussed earlier, models achieve surface consistency at the cost of cultural depth.

Although model performance is reported across three representative text types, we note that BLEU, CHRF and other n-gram–based metrics may respond differently to variations in text length, syntactic complexity and semantic density. Therefore, score comparisons across corpora, such as scientific abstracts, historical passages and literary lyrics, should be interpreted with caution, as metric sensitivity may be strongly corpus-dependent.

\section{Comparative Evaluation of Back-Translation Quality Across Multiple Samples and Platforms}

To ensure statistical significance in the evaluation of Chinese back-translation quality, this section uses a multi-sample corpus, \textit{CNKI-CHE-89}, which consists of 89 abstracts randomly selected from a larger dataset of 295 chemical abstracts crawled from the CNKI website. Five large language models (LLMs) were used for translation. BLEU metrics were computed using Jieba segmentation and two n-gram weighting schemes: 

\begin{itemize}
  \item \textbf{BLEU} (the n-gram weight distribution is: 0.5, 0.5, 0, 0) ,  motivated by modern Chinese word distribution statistics \cite{liu1986chinese}, where one-character words account for 56.7\%, two-character words for 39.65\% and others for 3.65\%.
  \item \textbf{BLEU-Unif} (the n-gram weight distribution is: 0.25, 0.25, 0.25, 0.25) ,  matching uniform weighting commonly adopted by LLMs like ChatGPT, Grok and Gemini.
\end{itemize}

\subsection{Experiment Results}
As shown in Table~\ref{tab:four_metrics I}, translation quality metrics varied across LLMs, revealing distinct patterns between reasoning-enabled and non-reasoning models. All translations and evaluations were conducted between March and April 2025. On average, BLEU scores reached 0.569 (SD = 0.112), while BLEU-Unif averaged 0.386 (SD = 0.133). CHRF, TER and semantic similarity scores were 0.838, 0.082 and 0.105, respectively. Notably, the wider variance observed in BLEU-Unif and semantic similarity suggests greater inconsistency in meaning preservation and model behavior across different systems.

\begin{table*}[ht]
    \centering
    \small
    \caption{Global Descriptive Statistics of Translation Metrics}
    \label{tab:global_metrics}
    \begin{tabular}{lccccc}
        \toprule
        Metric & BLEU & BLEU-Unif & CHRF & TER & Semantic Sim. \\
        \midrule
        Count & 1335 & 1335 & 1335 & 1335 & 1335 \\
        Mean & 0.569 & 0.386 & 0.838 & 0.082 & 0.105 \\
        Standard Deviation & 0.112 & 0.133 & 0.062 & 0.101 & 0.103 \\
        Min & 0.000 & 0.000 & 0.000 & 0.013 & 0.000 \\
        25\% & 0.494 & 0.296 & 0.803 & 0.053 & 0.039 \\
        50\% (Median) & 0.576 & 0.385 & 0.844 & 0.069 & 0.079 \\
        75\% & 0.650 & 0.477 & 0.879 & 0.087 & 0.135 \\
        Max & 0.931 & 0.925 & 0.969 & 1.000 & 0.691 \\
        \bottomrule
    \end{tabular}
    \label{tab:four_metrics I}
\end{table*}

\begin{table*}[htbp]
    \centering
    \caption{Descriptive statistics by model}
    \label{tab:transposed_model_stats}
    \begin{tabular}{llccccc}
        \toprule
        \textbf{Metric} & \textbf{Statistic} & 
        \textbf{\makecell{Claude\\Sonnet-3.7}} & 
        \textbf{\makecell{DeepSeek\\V3}} & 
        \textbf{\makecell{Gemini-2\\FTE}} & 
        \textbf{\makecell{Grok\\Beta}} & 
        \textbf{\makecell{OpenAI\\GPT-4.5}} \\
        \midrule
        \multirow{7}{*}{BLEU\tablefootnote{Higher is better. Measures precision of n-grams matching the reference.}} 
            & mean  & 0.5935 & \textbf{0.6033} & 0.5997 & 0.5223 & 0.5275 \\
            & std   & \textbf{0.1014} & 0.1023 & 0.1138 & 0.1073 & 0.1072 \\
            & min   & 0.2741 & \textbf{0.2770} & 0.0000 & 0.2227 & 0.2573 \\
            & 25\%  & 0.5285 & 0.5277 & \textbf{0.5338} & 0.4503 & 0.4578 \\
            & 50\%  & 0.5999 & 0.6059 & \textbf{0.6081} & 0.5313 & 0.5318 \\
            & 75\%  & 0.6689 & 0.6781 & \textbf{0.6797} & 0.6012 & 0.5931 \\
            & max   & 0.8272 & 0.8249 & 0.8348 & 0.7905 & \textbf{0.9309} \\
        \midrule

        \multirow{7}{*}{BLEU-Unif\tablefootnote{Higher is better. Measures precision of n-grams matching the reference.}} 
            & mean  & 0.4134 & \textbf{0.4283} & 0.4248 & 0.3311 & 0.3364 \\
            & std   & 0.1257 & 0.1298 & 0.1313 & \textbf{0.1191} & 0.1238 \\
            & min   & 0.0000 & 0.0000 & 0.0000 & 0.0000 & \textbf{0.0790} \\
            & 25\%  & 0.3335 & 0.3382 & \textbf{0.3498} & 0.2557 & 0.2515 \\
            & 50\%  & 0.4068 & 0.4185 & \textbf{0.4334} & 0.3285 & 0.3305 \\
            & 75\%  & 0.5036 & \textbf{0.5153} & 0.5060 & 0.4133 & 0.4046 \\
            & max   & 0.7449 & 0.7281 & 0.7509 & 0.6747 & \textbf{0.9254} \\
        \midrule
        
        \multirow{7}{*}{CHRF\tablefootnote{Higher is better. Measures character-level similarity.}} 
            & mean  & 0.8494 & 0.8494 & \textbf{0.8530} & 0.8127 & 0.8244 \\
            & std   & \textbf{0.0476} & 0.0515 & 0.0763 & 0.0637 & 0.0565 \\
            & min   & \textbf{0.6897} & 0.6889 & 0.0000 & 0.6238 & 0.6415 \\
            & 25\%  & 0.8178 & 0.8155 & \textbf{0.8269} & 0.7715 & 0.7843 \\
            & 50\%  & 0.8519 & 0.8462 & \textbf{0.8632} & 0.8168 & 0.8257 \\
            & 75\%  & 0.8830 & 0.8871 & \textbf{0.8932} & 0.8612 & 0.8653 \\
            & max   & 0.9608 & 0.9655 & \textbf{0.9688} & 0.9438 & 0.9455 \\
        \midrule
        \multirow{7}{*}{\makecell{Semantic\\Similarity}\tablefootnote{Higher is better. Measures semantic closeness between generated and reference text.}} 
            & mean  & 0.1172 & \textbf{0.1287} & 0.1144 & 0.0788 & 0.0879 \\
            & std   & 0.1038 & 0.1190 & 0.1020 & \textbf{0.0788} & 0.0969 \\
            & min   & 0.0000 & 0.0000 & 0.0000 & 0.0000 & 0.0000 \\
            & 25\%  & 0.0483 & \textbf{0.0533} & 0.0507 & 0.0267 & 0.0267 \\
            & 50\%  & 0.0887 & \textbf{0.1011} & 0.0887 & 0.0632 & 0.0674 \\
            & 75\%  & \textbf{0.1684} & 0.1664 & 0.1446 & 0.1097 & 0.1125 \\
            & max   & 0.4738 & \textbf{0.6586} & 0.5732 & 0.4409 & 0.6910 \\
            \midrule
        \multirow{7}{*}{TER\tablefootnote{Lower is better. Measures the number of edits needed to match the reference.}} 
            & mean  & 0.0819 & \textbf{0.0808} & \textbf{0.0808} & 0.0825 & 0.0824 \\
            & std   & 0.1014 & 0.1016 & 0.1015 & \textbf{0.1012} & 0.1016 \\
            & min   & 0.0160 & 0.0142 & \textbf{0.0133} & 0.0153 & 0.0163 \\
            & 25\%  & 0.0541 & 0.0522 & \textbf{0.0520} & 0.0535 & 0.0538 \\
            & 50\%  & 0.0700 & \textbf{0.0679} & 0.0690 & 0.0713 & 0.0693 \\
            & 75\%  & 0.0889 & 0.0870 & \textbf{0.0847} & 0.0908 & 0.0903 \\
            & max   & 1.0000 & 1.0000 & 1.0000 & 1.0000 & 1.0000 \\
        \bottomrule
    \end{tabular}
\end{table*}

\begin{figure*}[ht]
    \centering
    \includegraphics[width=0.95\textwidth]{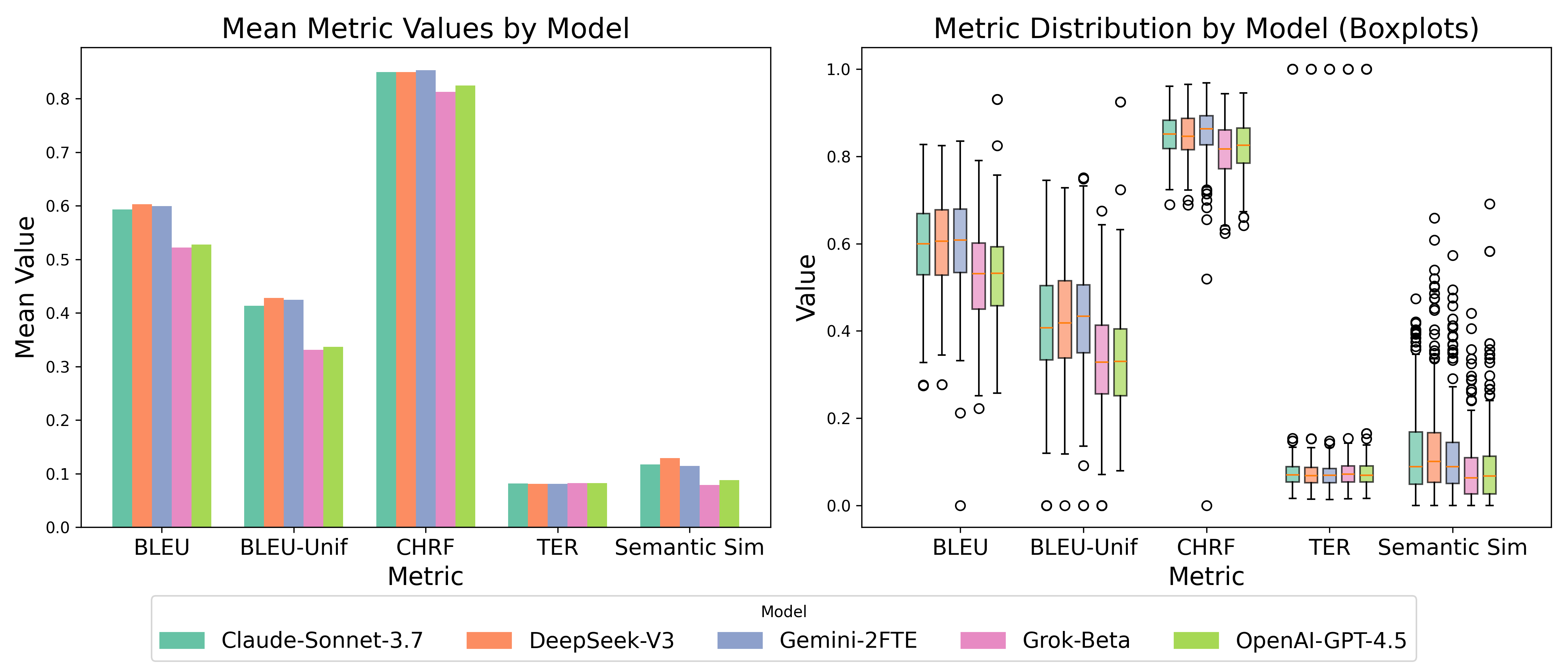}
    \caption{Comparison of translation metrics across models.}
    \label{metrics_by_model}
\end{figure*}

Table~\ref{tab:transposed_model_stats} details the scores by model. Reasoning-enabled models (Claude-Sonnet-3.7, Gemini-2FTE, Grok-Beta) present BLEU means ranging from 0.522 to 0.600 and BLEU-Unif means from 0.331 to 0.425, with CHRF scores between 0.81 and 0.85, indicating consistent surface-level translation quality. Among non-reasoning models, DeepSeek-V3 achieves the highest BLEU (0.603) and BLEU-Unif (0.428), as well as the highest Semantic Similarity (0.129), suggesting strong preservation of meaning. In contrast, OpenAI-GPT-4.5 reports lower BLEU (0.528) and BLEU-Unif (0.336), alongside a reduced Semantic Similarity (0.088), implying reduced semantic alignment despite high maximum BLEU-Unif peaks.

The boxplot analysis (Figure~\ref{metrics_by_model}) reveals the variability in BLEU, BLEU-Unif and Semantic Similarity scores, with outliers in the latter two indicating potential inconsistencies in translation quality. Compact distributions in BLEU, BLEU-Unif and CHRF suggest stable n-gram and character-level overlap, whereas higher dispersion in Semantic Similarity and BLEU-Unif reflects variability in meaning retention and lexical diversity. DeepSeek-V3 and Claude-Sonnet-3.7 exhibit more consistent performance, while OpenAI-GPT-4.5 and Grok-Beta display higher variability, indicating less reliable translation quality across abstracts.

Further examination of Semantic Similarity and BLEU-Unif reveals distinct model behaviors. DeepSeek-V3 achieves the highest means in both metrics but also shows increased dispersion, suggesting occasional semantic lapses or inconsistencies in lexical choices. Claude-Sonnet-3.7 yields similar averages with lower variability, indicating greater stability. Conversely, OpenAI-GPT-4.5 and Grok-Beta register lower scores and broader distributions, potentially due to misinterpretation of domain-specific terminology or artifacts from back-translation strategies. These findings underscore the importance of evaluating both lexical precision and semantic adequacy, especially in scientific abstracts where conceptual fidelity is essential.

In applied scenarios, such differences are particularly relevant for technical domains such as chemistry, where terminological precision and semantic clarity are crucial. Reasoning-enabled models tend to show greater consistency in structural metrics (BLEU, BLEU-Unif and CHRF), but the strong performance of DeepSeek-V3, a non-reasoning model, demonstrates effective semantic preservation, leading all models in BLEU, BLEU-Unif and Semantic Similarity.

Therefore, if semantic retention is prioritized, DeepSeek-V3 may offer practical advantages. In contrast, when structural consistency or domain-specific terminology must be preserved with minimal variation, reasoning-enabled models may provide more stable outputs. Ultimately, the choice of model depends on the trade-off between semantic accuracy and structural reliability for the specific use case.

Here, the term \emph{outperformance} is used strictly in relation to the quantitative metrics employed (BLEU, BLEU-Unif, CHRF, Semantic Similarity, TER). It does not imply comprehensive superiority in domains such as literary style, cultural fidelity or interpretative depth, which require more nuanced human evaluation.

\subsubsection{Correlation Between Translation Evaluation Metrics and Semantic Similarity}

\begin{figure*}[ht]
    \centering
    \includegraphics[width=0.9\textwidth]{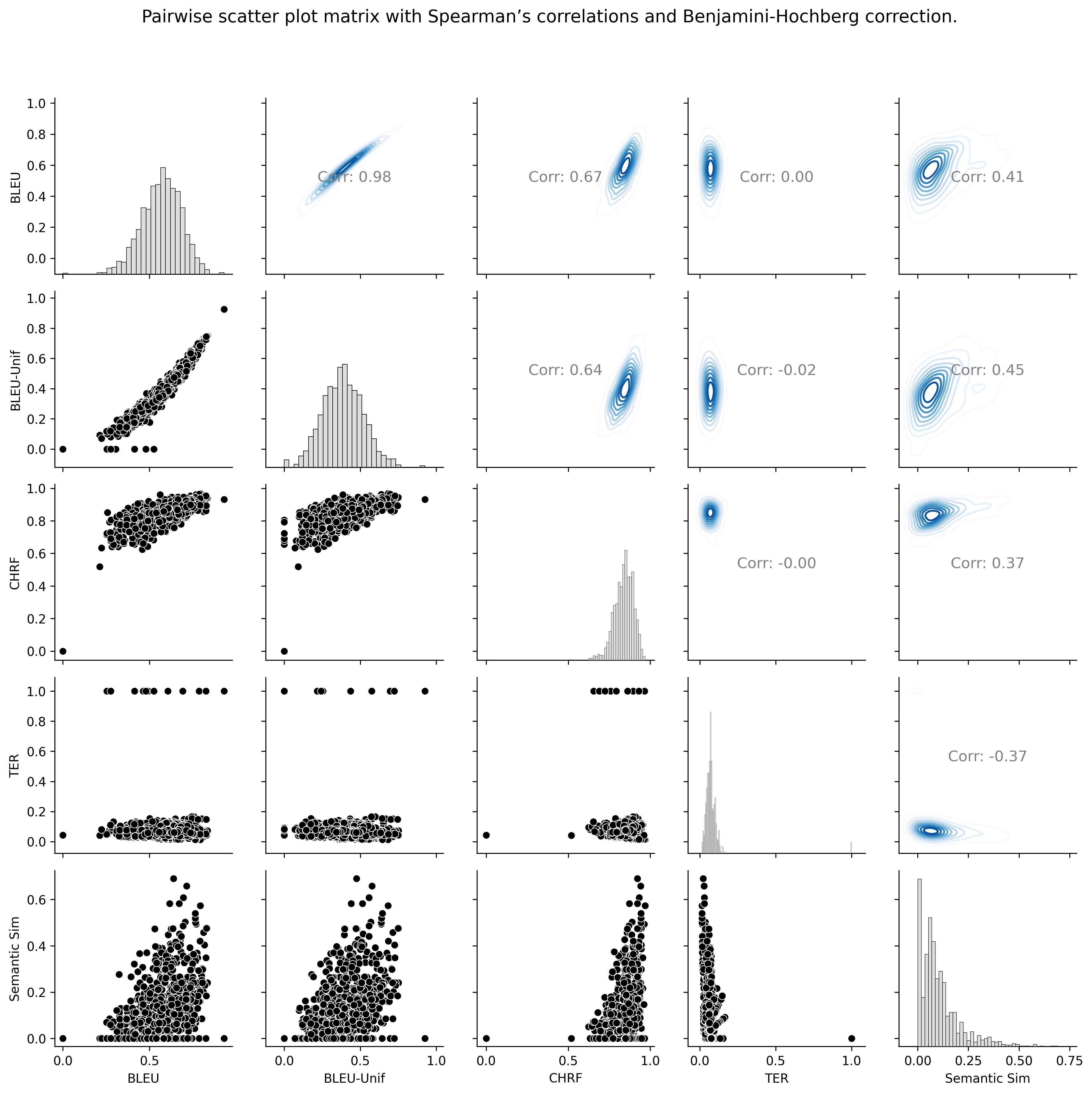}
    \caption{Pairwise scatter plot matrix with Spearman’s correlations and Benjamini-Hochberg correction.}
    \label{pairwise_correlation_models}
\end{figure*}

This analysis investigates the relationships among BLEU, BLEU-Unif, CHRF, TER and Semantic Similarity evaluation metrics, both in aggregate (Figure~\ref{metrics_by_model}) and across individual models (Figure~\ref{pairwise_correlation_models}), using Spearman correlation and Benjamini-Hochberg correction.

In the aggregated view, BLEU and CHRF present a correlation of 0.67, suggesting a moderately strong alignment in translation quality estimation. Across individual models, this relationship varies from 0.59 (Claude-Sonnet-3.7, Gemini-2FTE) to 0.70 (Grok-Beta), with Grok-Beta exhibiting the strongest alignment. BLEU-Unif shows consistently high correlation with BLEU across models (above 0.96), confirming that both metrics reflect similar surface-level lexical behavior, though BLEU-Unif incorporates a more balanced n-gram weighting scheme.

BLEU correlates moderately with Semantic Similarity (0.41 overall), with DeepSeek-V3 showing the strongest alignment (0.43), followed closely by Grok-Beta (0.40) and Gemini-2FTE (0.39). BLEU-Unif improves upon this slightly, reaching a higher correlation with Semantic Similarity across all models: 0.47 in DeepSeek-V3, 0.44 in Gemini-2FTE, 0.43 in Grok-Beta, 0.41 in Claude-Sonnet-3.7 and 0.34 in GPT-4.5. These values suggest that BLEU-Unif may be more sensitive to semantic preservation. However, we emphasize that these correlations reflect co-variation and do not imply causality. The observed alignment in models like DeepSeek-V3 may indicate consistent metric behavior, but further experimental control is needed to establish any direct explanatory role regarding translation quality.

CHRF maintains moderate correlations with Semantic Similarity, ranging from 0.30 (Grok-Beta) to 0.38 (DeepSeek-V3), generally lower than those observed for BLEU-Unif. This indicates that character-level overlap captures some semantic information, but less so than n-gram–based metrics with broader coverage like BLEU-Unif.

In contrast, TER behaves differently. It exhibits a negative overall correlation with Semantic Similarity (-0.37), with DeepSeek-V3 again showing the strongest inverse relationship (-0.43), followed by Claude-Sonnet-3.7 (-0.38) and Grok-Beta (-0.37). These results confirm that greater edit distance implies reduced semantic alignment. Moreover, TER shows near-zero correlation with BLEU (-0.01 to 0.06), BLEU-Unif (-0.04 to 0.03) and CHRF (–0.04 to 0.07), reinforcing its independence as a metric focused on edit operations rather than semantic or lexical similarity.

From a distributional perspective, BLEU, BLEU-Unif and CHRF deviate from normality (Shapiro–Wilk $p < 0.0001$), with BLEU showing slight right-skew, BLEU-Unif exhibiting wider tails and CHRF concentrating near its upper bound. TER values are tightly clustered at low levels, reflecting minimal required corrections. Semantic Similarity is skewed toward the lower end, suggesting that many translations struggle with semantic fidelity. Although the aggregate view provides an overview, the model-wise matrices better expose differences in dispersion, alignment and robustness of each metric.

Overall, DeepSeek-V3 presents the strongest internal coherence, with the highest BLEU-Unif–Semantic Similarity correlation (0.47), confirming its ability to retain meaning consistently. Gemini-2FTE follows closely, combining high BLEU-Unif alignment with moderate CHRF and Semantic Similarity scores. Grok-Beta exhibits the highest BLEU–CHRF correlation (0.70), but more limited semantic fidelity. Claude-Sonnet-3.7 offers balanced but moderate alignment across all metrics, while GPT-4.5 demonstrates the weakest semantic correlation in both BLEU (0.30) and BLEU-Unif (0.34), limiting its suitability in semantically demanding applications.

\subsubsection{Friedman Test and Dunn Post-Hoc Analysis}

The Friedman test, a non-parametric method for paired data, was used to assess differences between translation models based on average metric values. A p-value below 0.05 indicates significant differences, prompting a post-hoc Dunn test for pairwise model comparisons. The Benjamini-Hochberg correction was applied to manage false positives, in contrast to the more conservative Bonferroni method. Additionally, mean differences between statistically distinct model pairs were calculated to quantify the magnitude of those differences. This approach ensures robust and interpretable results, highlighting both the presence and practical relevance of observed differences.

The statistical analysis revealed significant differences across all evaluated metrics, BLEU, BLEU-Unif, CHRF and Semantic Similarity, with the exception of TER, as indicated by Friedman test p-values approaching zero. This confirms that at least one model differs significantly from the others in each metric, except for edit-distance–based performance.

For BLEU, which measures translation fidelity via weighted n-gram overlap, Claude-Sonnet-3.7, DeepSeek-V3 and Gemini-2FTE significantly outperformed Grok-Beta and OpenAI-GPT-4.5, with mean differences ranging from 0.0659 to 0.0810. These findings suggest that the latter models likely perform more reformulations during translation, reducing fidelity to the original text. A similar pattern was observed for BLEU-Unif, which distributes weights uniformly across 1- to 4-grams: the same three models outperformed Grok-Beta and GPT-4.5 with larger mean differences (0.0770 to 0.0972), reinforcing that BLEU-Unif captures additional translation variability. These higher deltas may reflect broader phrase-level divergence, such as that introduced by back-translation strategies.

The CHRF metric, which is sensitive to character-level changes and morphological variation, also showed significant differences between the same groups, though with smaller mean differences (all below 0.05). This suggests that Claude-Sonnet-3.7, DeepSeek-V3 and Gemini-2FTE preserve surface structures more consistently, whereas Grok-Beta and GPT-4.5 tend to introduce morphological variations that slightly reduce CHRF scores without major distortions.

Although the Friedman test indicated significance for TER, Dunn’s post-hoc test revealed no statistically distinguishable pairs. This discrepancy may result from the floor effect caused by low and concentrated TER values, which limits the sensitivity of the post-hoc comparisons. The results imply that despite minor variations, all models require similar levels of post-editing effort to match reference translations.

The Semantic Similarity metric, focused on meaning preservation, also revealed statistically significant differences. Again, Claude-Sonnet-3.7, DeepSeek-V3 and Gemini-2FTE outperformed Grok-Beta and GPT-4.5, with mean differences ranging from 0.0265 to 0.0499. These differences suggest that Grok-Beta and GPT-4.5 may deviate more frequently from the intended meaning, possibly due to hallucinations or overgeneralizations.

Taken together, Claude-Sonnet-3.7, DeepSeek-V3 and Gemini-2FTE emerge as more suitable for high-fidelity translation tasks, such as technical, scientific or legal documents, where precision and consistency are paramount. Conversely, Grok-Beta and OpenAI-GPT-4.5 may be preferable for creative or literary content, where semantic flexibility and stylistic variation are desirable. Given the absence of significant TER differences, the overall post-editing workload remains comparable across models. Therefore, model selection should be guided by the specific translation context, balancing structural stability, terminological accuracy and semantic preservation.

\begin{table}[h]
    \centering
    \caption{Summary of Statistical Differences Between Translation Models}
    \begin{tabular}{cc}
        \hline
        \textbf{Metric} & \textbf{Statistically Different Models and Mean Differences} \\
        \hline
        \multirow{6}{*}{BLEU} & Claude-Sonnet-3.7 vs Grok-Beta (0.0712) \\
                              & Claude-Sonnet-3.7 vs OpenAI-GPT-4.5 (0.0659) \\
                              & DeepSeek-V3 vs Grok-Beta (0.0810) \\
                              & DeepSeek-V3 vs OpenAI-GPT-4.5 (0.0757) \\
                              & Gemini-2FTE vs Grok-Beta (0.0774) \\
                              & Gemini-2FTE vs OpenAI-GPT-4.5 (0.0722) \\
        \hline
        \multirow{6}{*}{BLEU-Unif} & Claude-Sonnet-3.7 vs Grok-Beta (0.0823) \\
                                   & Claude-Sonnet-3.7 vs OpenAI-GPT-4.5 (0.0770) \\
                                   & DeepSeek-V3 vs Grok-Beta (0.0972) \\
                                   & DeepSeek-V3 vs OpenAI-GPT-4.5 (0.0919) \\
                                   & Gemini-2FTE vs Grok-Beta (0.0937) \\
                                   & Gemini-2FTE vs OpenAI-GPT-4.5 (0.0884) \\
        \hline
        \multirow{6}{*}{CHRF} & Claude-Sonnet-3.7 vs Grok-Beta (0.0367) \\
                              & Claude-Sonnet-3.7 vs OpenAI-GPT-4.5 (0.0249) \\
                              & DeepSeek-V3 vs Grok-Beta (0.0368) \\
                              & DeepSeek-V3 vs OpenAI-GPT-4.5 (0.0250) \\
                              & Gemini-2FTE vs Grok-Beta (0.0404) \\
                              & Gemini-2FTE vs OpenAI-GPT-4.5 (0.0286) \\
        \hline
        \multirow{6}{*}{Semantic Similarity} & Claude-Sonnet-3.7 vs Grok-Beta (0.0384) \\
                                             & Claude-Sonnet-3.7 vs OpenAI-GPT-4.5 (0.0294) \\
                                             & DeepSeek-V3 vs Grok-Beta (0.0499) \\
                                             & DeepSeek-V3 vs OpenAI-GPT-4.5 (0.0409) \\
                                             & Gemini-2FTE vs Grok-Beta (0.0356) \\
                                             & Gemini-2FTE vs OpenAI-GPT-4.5 (0.0265) \\
        \hline
        TER & No statistically significant differences between pairs \\
        \hline
    \end{tabular}
\end{table}

\subsubsection{Balancing Fidelity and Fluency - Ranking Translation Models}

Translation model evaluation ranked systems by BLEU, BLEU-Unif, CHRF, TER and Semantic Similarity. DeepSeek-V3 excelled with the highest BLEU (0.6033), BLEU-Unif (0.4283) and Semantic Similarity (0.1287), indicating strong lexical fidelity, balanced n-gram coverage and robust meaning preservation. It also reported one of the lowest TER scores (0.0808), reflecting low post-editing effort. Despite lacking reasoning capabilities, DeepSeek-V3 outperformed all other models, challenging the assumption that reasoning is a prerequisite for high-quality translation.

Gemini-2FTE ranked second, with a competitive BLEU (0.5997) and BLEU-Unif (0.4248) and the highest CHRF score (0.8530), indicating precise character-level alignment. Its TER (0.0808) matched DeepSeek-V3, while Semantic Similarity (0.1144) was slightly lower, suggesting a minor trade-off in semantic retention. Like DeepSeek-V3, Gemini-2FTE achieved top-tier results without relying on explicit reasoning mechanisms, highlighting the potential of non-reasoning models when well-optimized.

Claude-Sonnet-3.7 placed third with BLEU (0.5935), BLEU-Unif (0.4134) and CHRF (0.8494) scores close to the leaders. Its TER (0.0819) was slightly higher, indicating marginally increased editing effort. However, its Semantic Similarity (0.1172) surpassed Gemini-2FTE, reflecting strong meaning preservation. As a reasoning-enabled model, its performance aligns with expectations, though not surpassing the best non-reasoning models.

OpenAI-GPT-4.5 ranked fourth, with lower BLEU (0.5275), BLEU-Unif (0.3364) and CHRF (0.8244) scores, pointing to increased paraphrasing and reduced surface similarity. Its Semantic Similarity (0.0879) was considerably lower, suggesting notable semantic drift. Although TER (0.0824) was comparable to others, its overall translation quality was less consistent. As a reasoning-capable model, GPT-4.5 underperformed relative to non-reasoning models, emphasizing that reasoning alone does not guarantee superior translation.

Grok-Beta ranked last, with the lowest BLEU (0.5223), BLEU-Unif (0.3311) and CHRF (0.8127) scores, indicating poor n-gram and character overlap. It also recorded the highest TER (0.0825) and the lowest Semantic Similarity (0.0788), highlighting considerable meaning loss and suggesting higher post-editing demand. This model appears less suitable for accuracy-sensitive translation tasks.

In summary, DeepSeek-V3 demonstrated that high-quality translation can be achieved without reasoning, leading in BLEU, BLEU-Unif and Semantic Similarity. Claude-Sonnet-3.7 and Gemini-2FTE provided balanced performance, while OpenAI-GPT-4.5 and Grok-Beta lagged in semantic and lexical fidelity. These findings indicate that reasoning is not a prerequisite for translation excellence and that well-designed non-reasoning models can outperform more complex architectures in practical scenarios.

\color{black}

\section{Discussion (1): Analysis of Anomalies in Back-Translation}

Figure~\ref{metrics_by_model} not only displays the boxplot for major evaluation metrics but also reveals the presence of outliers. This section explores these anomalies to analyze specific phenomena observed in back-translation using large language models (LLMs).

\subsection{SOTA Performance in ZH$\rightarrow$EN and EN$\rightarrow$ZH Translation}

Zhu et al. (2025) \cite{zhu2025overcoming} reported state-of-the-art (SOTA) performance across 23 LLMs. Their study showed that for Chinese-to-English (ZH$\rightarrow$EN) translation, GPT-4.0 achieved the highest quality with BLEU $\leq 27.26$, notably lower than BLEU scores for translations from German, Russian or Czech into English. For English-to-Chinese (EN$\rightarrow$ZH) translation, GPT-3.5-T led the results with BLEU  $\leq 44.63$. Although their evaluation included a wide range of MoE-based LLMs, it did not cover more recent platforms such as Grok, DeepSeek or Claude. Still, the study reflects a current reality: Chinese-to-English remains weaker in machine translation performance compared to other language pairs.

In contrast, BLEU scores in our single- and multi-sample experiments often surpass Zhu’s benchmarks. Possible explanations include:
\begin{itemize}
  \item \textbf{Domain specificity:} Our corpus focuses on chemical terms and named entities, which may yield higher BLEU scores.
  \item \textbf{Evaluation method:} Back-translation filters unstable outputs, potentially inflating BLEU.
  \item \textbf{Model generation gap:} Newer LLMs (e.g., GPT 4.5, DeepSeek V3) likely outperform older models used in Zhu et al. \cite{zhu2025overcoming}.
  \item \textbf{Metric variation:} Differences in tokenization, reference alignment and n-gram schemes affect BLEU values.
\end{itemize}

\subsection{Occasional Traditional Chinese Output}

An anomaly was observed in sample CNKI-CHE-89-28, where Gemini 2.0 generated traditional Chinese text in its second round of back-translation, causing sharp drops in evaluation metrics. Table~\ref{tab:gemini_tradition} shows the BLEU, CHRF, TER and semantic similarity across the three trials.

\begin{table}[htbp]
\centering
\caption{Gemini 2.0 Back-Translation Scores for CNKI-CHE-89-28 (Simplified vs. Traditional Chinese)}
\label{tab:gemini_tradition}
\begin{tabular}{l|c|c|c|c|c}
\hline
\textbf{Round} & BLEU & BLEU-Unif & CHRF & TER & Semantic Sim. \\
\hline
First  & 0.8119 & 0.7133 & 0.8571 & 0.0329 & 0.4576 \\
Second & 0.2123 & 0.0914 & 0.5194 & 0.0417 & 0.0000 \\
Third  & 0.6778 & 0.4853 & 0.8442 & 0.0347 & 0.4275 \\
\hline
Mean   & 0.5673 & 0.4300 & 0.7403 & 0.0364 & 0.2950 \\
SD     & 0.0948 & 0.1611 & 0.0092 & 0.0095 & 0.0213 \\
\hline
\end{tabular}
\end{table}

In the second round, scores across all five metrics dropped dramatically. Upon inspection, the back-translated text was found to be in traditional Chinese, e.g., ``\begin{CJK}{UTF8}{bsmi}元素觀點：物質由元素組成\end{CJK} (Elemental view: Matter is composed of elements)'', diverging from the original simplified Chinese. This discrepancy explains the statistical variance and exemplifies a rare but critical failure point: LLMs must support both simplified and traditional Chinese language systems. Clear task instructions specifying the desired Chinese variant (e.g., Simplified for Mainland China) are crucial.

Additionally, Gemini 2.0 failed to generate any result for sample CNKI-CHE-89-79 during one round, indicating possible system refusal or inference failure. Though such anomalies are rare (1 in 1,335), they meaningfully affect aggregated results and highlight the complexities of Chinese NLP.

\subsection{Verbatim Back-Translation (Mirrored Output)}
Verbatim back-translation refers to a phenomenon where the model outputs text nearly identical to the original input, $ZHy \approx ZHx$, bypassing actual translation, see Figure \ref{imagery_vs_selfawareness}. This inflates automatic metrics like BLEU, as the output matches the reference but lacks true cross-lingual processing.

As shown in Table \ref{tab:xue_dejiong}, DeepSeek’s translation of Xue Dejiong’s Dilemma nearly mirrors the source sentence. Given the complexity of the content, such high BLEU scores likely result from retrieval or memorization, rather than proper semantic transformation. This mirrored output is more prevalent in LLM-based back-translation than in NMT systems.

Further testing with Mistral in a ZH$\rightarrow$FR$\rightarrow$ZH pipeline also yielded instances of verbatim copying, reinforcing that this is a cross-lingual issue. However, in the second DeepSeek translation round, the output changed to:

\begin{quote}
\begin{CJK*}{UTF8}{gbsn}满纸都是胡言乱语，仿佛外文书籍。即使有谪仙李太白（李白）这样杰出的人物，恐怕也难以理解。\end{CJK*}
\end{quote}

indicating the absence of the mirrored effect. This variability reflects the non-deterministic nature of LLM behavior under different conditions.

\subsection{Extreme Performance Variance in CNKI-CHE-89-48}

This case involves a short abstract \cite{ma2024cfm} from the CNKI-CHE-89 corpus:

\begin{quote}
\begin{CJK*}{UTF8}{gbsn}探讨了氨基寡糖素与化学杀菌剂复配对大白菜霜霉病的防治效果。\end{CJK*}
\end{quote}

(The study explored the efficacy of amino-oligosaccharins combined with chemical fungicides for controlling downy mildew in Chinese cabbage.)

In three back-translation runs by GPT-4.5:

\begin{itemize}
    \item GPT-4.5 achieved BLEU = 0.9309 (twice), with a mean of 0.8957.
    \item Back-translated output was:  
    \begin{quote}
    \begin{CJK*}{UTF8}{gbsn}研究了氨基寡糖素与化学杀菌剂复配对大白菜霜霉病的防治效果。\end{CJK*}
    \end{quote}
    nearly identical to the original, $ZHy \approx ZHx$.
    \item When the same English version (EN) was manually translated back by GPT-4, the result was:  
    \begin{quote}
    \begin{CJK*}{UTF8}{gbsn}研究了氨基寡糖与化学杀菌剂联合对白菜霜霉病的防控效果。\end{CJK*}
    \end{quote}
    This version scored BLEU = 0.7566 (char-level), slightly lower due to lexical differences:  
    \begin{CJK*}{UTF8}{gbsn}``探讨''\end{CJK*} vs. \begin{CJK*}{UTF8}{gbsn}``研究''\end{CJK*}, \begin{CJK*}{UTF8}{gbsn}``复配''\end{CJK*} vs. \begin{CJK*}{UTF8}{gbsn}``联合''\end{CJK*}, \begin{CJK*}{UTF8}{gbsn}``防治''\end{CJK*} vs. \begin{CJK*}{UTF8}{gbsn}``防控''\end{CJK*}.
\end{itemize}

Other platforms performed far worse: Grok BLEU mean: 0.4030 (min 0.2535) and Claude BLEU mean: 0.4264 (min 0.2761).

These findings suggest that GPT-4.5 may possess implicit task quasi-self-awareness, recognizing the back-translation loop and generating output that maximizes BLEU, potentially through memorization. While this increases automatic scores, it raises questions about actual translation fidelity.

\section{Discussion (2): Human--Machine Differences and the Paradox of Poetic Intent}

This section explores two core concepts introduced in this study: the Paradox of Poetic Intent in human–machine back-translation divergence and verbatim back-translation as a reflection of quasi-self-awareness in LLMs. These ideas highlight underexamined phenomena in language translation and invite deeper inquiry. Interestingly, while such insights originate from human reflection, they often resonate with LLM behavior, suggesting that human creativity, paired with model refinement, can drive conceptual innovation.

\subsection{Causes of Verbatim Back-Translation in LLMs}

LLMs may sometimes ``realize'' they are performing a back-translation task, leading to direct reproduction of the original input---termed \textit{verbatim back-translation}. Several factors may contribute:

\begin{enumerate}
  \item \textbf{Alignment Bias}: LLMs prioritize fidelity. When detecting high-confidence bilingual pairs, they may replicate the source directly.
  \item \textbf{Implicit Retrieval}: LLMs predict based on contextual likelihood. If a source-target pair is well-established, especially in short or technical texts, retrieval outweighs transformation.
  \item \textbf{Task Pattern Recognition}: In round-trip scenarios ($X \rightarrow Y \rightarrow X$), the model may identify the task and return $X$.
  \item \textbf{Training Data Bias}: Exposure to standardized corpora increases preference for fixed expressions.
\end{enumerate}

This phenomenon signals a caution for MT evaluation: overreliance on automatic metrics (e.g., BLEU) may overlook deeper semantic fidelity. Improving model design, curating training data and employing diverse evaluation approaches can help reduce such effects and better reflect true translation competence.

\subsection{The Limits of Back-Translation: Machine vs. Human Translation}

Unlike humans, who adapt translation creatively \cite{wang2009translation}, LLMs tend to match high-probability outputs. Literary back-translations often distort nuance because:

\begin{itemize}
  \item Human translators reframe meaning within cultural norms.
  \item Expression varies based on interpretation and context.
  \item Machines lean toward memorized equivalence.
\end{itemize}

LLMs, driven by statistical prediction, tend toward probabilistic matching rather than creativity. Hence, they exhibit mirrored outputs more frequently than humans.

As shown in earlier sections, DeepSeek-V3 and DeepThinkR1 demonstrate an ability to ``recognize" complex structures in the lyrics of Hua Yao due to:
\begin{itemize}
\item Associative learning from corpus-specific stylistics (e.g., temporal-spatial shifts, dialogue forms);
\item Pattern recognition in classical imagery (e.g., ``brown robe,'' ``yellow sash'').
\end{itemize}

Yet such recognition stems from probabilistic textual associations, not genuine literary understanding in LLMs. For instance, the model fails to grasp that ``Quanting–Hangzhou–Yuhang'' is a historical-geographic metaphor with the time evolution. This distinction underscores a core divergence between machine and human translation. Genuine understanding remains the threshold for Artificial General Intelligence (AGI).

\subsection{The Paradox of Poetic Intent and Quasi-Self-Awareness}

In Chinese back-translation, especially with metaphor-rich texts or scientific terms, models struggle with imagery preservation and temporal structure---the \textit{Paradox of Poetic Intent}. However, LLMs often output verbatim back-translations, revealing an emergent \textit{Quasi-self-awareness} driven by massive data exposure.

For instance, the ``Hua Yao Red and Yellow" corpus, a metaphor encompassing rich, poignant stories, historical responses, religious traditions and decorations, is often oversimplified into a literal description. These distortions arise from cross-linguistic challenges and the difficulty of preserving the nuanced, high-context nature of Chinese expression.

However, unlike rule-based or statistical systems, LLMs show a unique ability to mirror the original text, at times bypassing actual translation. This is not merely a failure, but a reflection of emergent quasi-self-awareness: the model's ability to recognize and reconstruct source patterns due to its massive data exposure.

Such behavior suggests that verbatim back-translation may reflect intelligent alignment, not just memorization. For instance, when DeepSeek-V3 outputs \begin{CJK*}{UTF8}{gbsn}尘凡儿缠我谤我笑我白了头\end{CJK*} (The mundane world entangles me, slanders me, mocks me, until my hair turns white.) without transformation, it reveals an instinct to reassemble familiar linguistic imagery   a step toward human-like intuition.

This coexistence of quasi-self-awareness and the Paradox of Poetic Intent provides new ground for AGI research. LLMs demonstrate not only language generation but also semantic reconstruction, cultural sensitivity and temporal reasoning, key components of advanced intelligence.

Thus, rather than treating such behavior as failure, we propose that it signifies a new frontier: using paradoxes in poetic and metaphorical back-translation as lenses to reveal, challenge and expand LLM capacities. This offers pathways for:

\begin{itemize}
  \item Enhanced CNLP (e.g., culturally aware fine-tuning);
  \item Human–AI collaborative translation;
  \item Moving closer to AGI by learning how LLMs simulate (or fail to simulate) deep semantics.
\end{itemize}

\subsection{Insights and Optimization Directions}
Comparing human translation to LLM behavior, we observe that human translators rarely fall into the trap of mirrored reproduction. This suggests future directions for machine translation optimization:

\begin{itemize}
  \item Enhanced context modeling: Equipping LLMs with deeper understanding of discourse and narrative.
  \item Human-in-the-loop training: Incorporating diverse annotations to enrich stylistic variance.
  \item Reducing overfit to training data: By adapting sampling strategies and lowering reliance on surface-level matching.
\end{itemize}

Human translation is inherently bounded, contextual and creative \cite{wang2009translation}. LLMs, driven by large-scale pattern matching, often overfit and exhibit ``awareness" through extreme alignment. In this light, back-translation is not merely a performance artifact, it reflects how LLMs represent knowledge and simulate memory.

Understanding these behaviors helps shift our view of LLMs: from deterministic generators to probabilistic agents capable of reflective behavior, inviting deeper exploration into their role in translation, meaning and human–machine symbiosis.

\section{Conclusions}
This study demonstrates that large language models (LLMs) have shown significant potential in replacing traditional tools for Chinese reverse translation, particularly achieving breakthroughs in processing scientific and technical texts. However, when it comes to tasks involving literary style or interdisciplinary contexts, LLMs still face dual challenges of semantic understanding and stylistic fidelity. These scenarios call for the incorporation of expert knowledge and multi-level evaluation strategies. Future research may explore hybrid strategies combining ``LLMs + few-shot fine-tuning'', balancing general language capabilities with adaptability to specialized domains.

Academic contributions of this research include:

\begin{itemize}
\item \textit{Proposed the ``Paradox of Poetic Intent'' for the first time.} This study highlights how LLMs, in pursuit of linguistic alignment and literal consistency, often sacrifice the literary aura and deeper meanings of the original text. This paradox offers a novel perspective for evaluating the cultural fidelity of generated text and fills a gap in Chinese Natural Language Processing (CNLP) between literary and technical translation evaluation. 
\item \textit{Developed a BLEU evaluation method tailored to Chinese linguistic features.} Building upon the standard BLEU framework, this study integrates Jieba word segmentation, term frequency weighting and n-gram matching to propose an improved BLEU method better suited for Chinese. A comparative analysis with traditional character-level BLEU is provided, offering a more refined quantitative tool for Chinese translation quality assessment.
\item \textit{Constructed a reverse translation evaluation framework and conducted statistical validation.} A Chinese-English-Chinese reverse translation framework was designed, utilizing statistical methods such as Friedman tests alongside BLEU, CHRF, TER and semantic similarity metrics. This systematic comparison between LLMs and commercial translation systems across multiple text samples offers a reproducible and scalable evaluation paradigm.
\item \textit{Revealed essential differences in translation behavior between LLMs and traditional tools.} The study identifies behavioral distinctions through the analysis of anomalies in experimental results. Traditional tools still exhibit strengths in handling rule-based texts (e.g., Sogou achieving BLEU scores up to 0.57 in some tasks), while LLMs demonstrate traits such as verbatim back-translation and self-awareness. The latter implies the model's tendency to adjust expression for fluency, which may lead to semantic drift.
\end{itemize}

Implications for CNLP Research:

\begin{itemize}
\item \textit{Promoting culturally sensitive model development.} While CNLP has made progress in technical domains, understanding of imagery, symbolism and metaphor remains limited. It is recommended to incorporate cultural meta-information during model training to shift from ``literal translation'' to ``poetic conveyance.'' For example, integrating Two-Step RAG architectures may enhance the model's ability to recognize polysemous contexts.
\item \textit{Improving evaluation systems for Chinese translation.} Given the limitations of English-centric BLEU metrics in reflecting Chinese translation quality, this study proposes a segmentation-based, multi-metric evaluation system. This provides a more interpretable and informative tool for assessing Chinese-generated text.
\item \textit{Uncovering structural issues in LLM translation mechanisms.} Features such as self-awareness and the Paradox of Poetic Intent are uniquely observed in LLM-based Chinese translation, especially in literary texts. Future work may consider reinforcement learning (e.g., RLHF) or iterative alignment strategies to guide models toward preserving poetic intent and semantic integrity in generation tasks.
\item \textit{Fostering interdisciplinary integration between technology and the humanities.} Translation is not merely linguistic transfer, it is cultural interpretation and reconstruction. The example of the ``Xue Dejiong Dilemma'' illustrates the limitations of technology in handling cultural nuances and highlights the need for tighter integration between CNLP, translation studies and literary theory to build culturally enriched AI language systems.
\end{itemize}

In summary, this study has explored the nuanced behaviors of large language models (LLMs) through the lens of verbatim back-translation, quasi-self-awareness and the cultural-semantic paradoxes they encounter in poetic and lyrical texts. While our analysis reveals certain limitations in current LLMs when handling deep cultural semantics and temporal metaphors, we recognize that such insights are made possible only through the foundational capabilities established by models like the GPT series. The study not only offers a systematic evaluation of current LLM performance in Chinese translation but also proposes conceptual breakthroughs (e.g., the Paradox of Poetic Intent) and methodological innovations (e.g., back-translation with multi-dimensional evaluation).

Back-translation, as a ``touchstone'' for language intelligence, contributes directly to advances in machine translation, content generation and cross-lingual retrieval. Building upon bidirectional Chinese–English translation, the framework introduced in this work can be further extended to other linguistically rich systems such as Japanese, paving the way for more nuanced human–LLM interaction and the advancement of Artificial General Intelligence (AGI).

%%%%%%%%%%%%%%%%%%%%%%%%%%%%%%%%%%%%%%%%%%%%%%%%%%%%%%%%%%%%%%

\section{ACKNOWLEDGMENT}
Sincere thanks to all the friends and colleagues who offered encouragement and insights throughout this research. Special thanks to the support of various large language models, translation and information platforms, including ChatGPT, Claude, CNKI, DeepSeek, Gemini, Grok, Mistral, QwenVL, as well as Google Translate, Baidu Translate and Sogou Translate.

\bibliographystyle{unsrtnat}
%\bibliography{references}

\end{document}